\documentclass[a4paper,twoside]{article}

\usepackage{epsfig}
\usepackage{subcaption}
\usepackage{calc}
\usepackage{amssymb}
\usepackage{amstext}
\usepackage{amsmath}
\usepackage{amsthm}
\usepackage{multicol}
\usepackage{pslatex}
\usepackage{apalike}
\usepackage{algorithm2e}
\usepackage[bottom]{footmisc}

\usepackage{url}
\usepackage{multirow}
\usepackage{comment}

\usepackage{SCITEPRESS}     
\usepackage{orcidlink}

\begin{document}

\title{Improving Geometric Consistency for 360-Degree Neural Radiance Fields in Indoor Scenarios}


\author{
    \authorname{Iryna Repinetska\sup{2},
                Anna Hilsmann \sup{1}\orcidlink{0000-0002-2086-0951}, 
                Peter Eisert\sup{1},\sup{2}\orcidlink{0000-0001-8378-4805}}
\affiliation{\sup{1}Fraunhofer Institute for Telecommunications,
Heinrich Hertz Institute, Berlin, Germany}
\affiliation{\sup{2}Department of Computer Science, Humboldt University, Berlin, Germany}
\email{
repineti@hu-berlin.de,
anna.hilsmann@hhi.fraunhofer.de,
peter.eisert@hu-berlin.de
}
}


\abstract{
Photo-realistic rendering and novel view synthesis play a crucial role in human-computer interaction tasks, from gaming to path planning. Neural Radiance Fields (NeRFs) model scenes as continuous volumetric functions and achieve remarkable rendering quality. However, NeRFs often struggle in large, low-textured areas, producing cloudy artifacts known as ''floaters'' that reduce scene realism, especially in indoor environments with featureless architectural surfaces like walls, ceilings, and floors.
To overcome this limitation, prior work has integrated geometric constraints into the NeRF pipeline, typically leveraging depth information derived from Structure from Motion or Multi-View Stereo. Yet, conventional RGB-feature correspondence methods face challenges in accurately estimating depth in textureless regions, leading to unreliable constraints. This challenge is further complicated in 360-degree ''inside-out'' views, where sparse visual overlap between adjacent images further hinders depth estimation.
In order to address these issues, we propose an efficient and robust method for computing dense depth priors, specifically tailored for large low-textured architectural surfaces in indoor environments. We introduce a novel depth loss function to enhance rendering quality in these challenging, low-feature regions, while complementary depth-patch regularization further refines depth consistency across other areas. Experiments with Instant-NGP on two synthetic 360-degree indoor scenes demonstrate improved visual fidelity with our method compared to standard photometric loss and Mean Squared Error depth supervision.
}

\onecolumn \maketitle \normalsize \setcounter{footnote}{0} \vfill

\section{\uppercase{Introduction}}
\label{sec:introduction}

Neural Radiance Fields (NeRFs) provide a novel solution to a fundamental challenge in computer vision: generating new views from a set of posed 2D images \cite{arandjelovic2021nerf}. By modeling a scene as a continuous volumetric function and encoding it into the weights of a neural network, NeRFs achieve a remarkable balance between geometry and appearance representation
\cite{mildenhall2021nerf}. This technology offers substantial benefits across applications such as virtual reality, augmented reality, and robotics, where high-fidelity visualization is critical. 

While NeRFs produce realistic renderings across diverse settings, indoor environments with large, low-textured surfaces---such as walls, floors, and ceilings---present unique challenges. These areas often lack distinctive visual features, which significantly hinders NeRF's ability to accurately reconstruct the scene, leading to potential inaccuracies that compromise the quality of the final render \cite{wang2022neuris}. This often leads to undesired artifacts in the rendered views, with one of the most common being ''floaters'' \cite{roessle2022dense}. They appear as cloudy, erroneous, detached elements within the scene, significantly degrading the visual quality and realism of the generated views. Their occurrence is closely tied to irregularities in density distribution, stemming from inaccurate geometric estimates during color-driven optimization \cite{roessle2022dense}. Imposing geometric constraints through depth supervision mitigates these issues, typically involving the comparison of rendered depth with ground truth data during the training process \cite{deng2022depth}. However, acquiring accurate depth priors is an inherently challenging task, as most depth estimation methods rely on visual cues such as texture, edges, and shading to determine depth, often leading to inaccuracies in featureless areas \cite{gasperini2023robust}---a common characteristic of many views in indoor datasets. Additionally, capturing views with a 360-degree camera further complicates the task. Since the ''inside-out'' viewing direction results in sparse visual overlap between adjacent images, it is harder to align features across views \cite{chen2023structnerf}.

To overcome these challenges, we introduce an efficient method for extracting dense depth priors specifically for large planar architectural surfaces in indoor spaces, such as ceilings, walls, and floors, which are particularly susceptible to floaters. 
Our approach is tailored to indoor environments, requiring basic conditions that are easily met in typical settings, such as aligning the Z-axis with the floor plane normal. We assume the scene to be captured by a 360-degree camera which efficiently scans the entire rooms while being moved through the scene. Mounted on a tripod or stand, it enables straightforward estimation of the ground plane. Additionally, we assume that the room's height is known or can be measured, which is generally true for most indoor settings. Our method is also supported by semantic segmentation information of the image data, providing class labels for wall, floor, and ceiling. Given the advanced state of current semantic segmentation techniques, numerous pre-trained models are available that can generate segmentation masks for these classes without requiring a computationally intensive training process \cite{chen2017rethinking}, \cite{ronneberger2015u}, \cite{badrinarayanan2017segnet}.

Recognizing that architectural surfaces delineate the boundaries of an indoor scene, we introduce a loss function that encourages the alignment of a ray’s termination with these boundary surfaces---walls, floor, and ceiling. This function also promotes the correct distribution of volumetric densities along the ray, ensuring that the regions the ray passes through before hitting a boundary represent empty space, while density increases sharply at the boundary surfaces.

To further address flawed density distribution in other areas, we implement a patch-based depth regularization method that utilizes bilateral or joint bilateral filtering to
smooth out depth inconsistencies while preserving edge information.

To evaluate our approach, we created two synthetic 360-degree indoor scenes. Rather than relying on stitched panoramic views, we propose an unconventional method that uses a series of unstitched views, facilitating precise estimation of both extrinsic and intrinsic camera parameters---critical for NeRFs pipelines---and avoiding the geometric distortions introduced by the typical stitching process. Additionally, we assume the 360-degree camera is mounted on a movable stand, enabling efficient capture of an entire room and supporting dense depth estimation of architectural surfaces.

Our results, demonstrated on a 360-degree indoor dataset with Instant-NGP, show that incorporating depth supervision with our planar architectural depth priors improves visual quality compared to methods that rely solely on photometric loss. Moreover, our proposed depth loss for boundary surfaces outperforms Mean Squared Error (MSE) loss on both datasets, yielding superior visual coherence. Additionally, integrating our patch-based depth regularization techniques further refines results, enhancing depth consistency across the scene. Last but not least, training with depth supervision using our depth priors accelerates the process, further enhancing the efficiency of our approach.

In summary, the main contributions of this work are as follows:
\begin{itemize}
    \item The generation of a synthetic 360-degree indoor dataset, comprising two distinct scenes, which we intend to make publicly available to support future research.
    \item The design of an algorithm for producing dense depth priors on planar architectural surfaces, such as walls, ceilings, and floors.
    \item The formulation of a new depth loss function tailored for these planar boundary surfaces.
    \item The development of a patch-based depth regularization technique, incorporating bilateral and joint bilateral filters.
\end{itemize}

\section{\uppercase{Related Work}}
\label{sec:related_work}
Research to enhance NeRFs rendering quality has led to various \textit{depth regularization} and \textit{depth supervision} methods aimed at improving rendering quality by refining the scene's geometry.

\textit{Implicit regularization} approaches leverage pre-trained models to encode geometry and appearance priors. For instance, Pixel-NeRF \cite{yu2021pixelnerf} directly integrates features from a convolutional neural network (CNN) trained on multiple scenes to condition the NeRF model, while DietNeRF \cite{jain2021dietnerf} incorporates a regularization term in its loss function to enforce consistency between high-level features across both known and novel views. However, these regularization methods often struggle when applied directly to indoor datasets due to domain gaps, as the CNNs are typically pre-trained on ImageNet \cite{deng2009imagenet}, which predominantly features natural images. Bridging this gap can be resource-intensive and may require additional fine-tuning \cite{chen2023structnerf}.

\textit{Explicit regularization} methods specifically target high-frequency artifacts by smoothing inconsistencies between adjacent regions. RegNeRF \cite{niemeyer2022regnerf}, for example, enforces similarity constraints on neighboring pixel patches, while InfoNeRF \cite{kim2022infonerf} minimizes a ray entropy model to maintain consistent ray densities across views.

Although regularization techniques can enhance rendering quality to some degree, their overall impact remains limited \cite{chen2023structnerf}. In contrast, \textit{depth supervision} addresses sparse scenarios and regions with less prominent visual features by providing a stronger optimization signal through an additional depth constraint that leverages depth priors and ensures consistency between rendered and ground truth depth \cite{rabby2023beyondpixels}. For instance, DS-NeRF \cite{deng2022depth} and Urban-NeRF \cite{rematas2022urban} incorporate a depth loss that adjusts the predicted depth to match available sparse depth data.

In the context of indoor scene synthesis, notable research efforts such as Dense Depth Priors \cite{roessle2022dense} and NerfingMVS \cite{wei2021nerfingmvs} have proposed methods to enhance NeRF performance by transforming sparse data points—typically a byproduct of the Structure from Motion preprocessing step used for estimating camera poses—into dense depth maps using a monocular depth completion model. In the first approach, these dense depth priors are leveraged to guide the NeRF optimization process, effectively accounting for uncertainty in depth estimation while minimizing the error between predicted and true depth values \cite{roessle2022dense}. NerfingMVS \cite{wei2021nerfingmvs} builds on this by calculating loss through comparisons between rendered depth and learned depth priors, incorporating confidence maps to weigh the reliability of the depth estimates. These supervision strategies generally yield superior results compared to those relying solely on sparse depth points \cite{wang2023digging}. However, their limitation lies in a lack of view consistency, as each view is processed individually during the depth completion step. StructNeRF \cite{chen2023structnerf} addresses this by incorporating photometric consistency, comparing source images with their warped counterparts from other viewpoints in visually rich regions. To handle non-textured areas, it introduces a regularization loss that enforces planar consistency, encouraging points within regions identified by planar segmentation masks to lie on a single plane. This approach helps maintain multi-view consistency, though the warping process significantly increases computational cost \cite{wang2023anisotropic}. Notably, methods that utilize depth supervision struggle in areas with low visual features, either because they inherit limitations from Structure from Motion or Multi-View Stereo depth estimates, or, as in the case of StructNeRF, rely on warping for photometric consistency.

Research on 360-degree panorama NeRF-based view synthesis, similar to the pinhole camera model, widely applies additional depth supervision for optimization \cite{gu2022omni}, \cite{wang2023perf}, \cite{kulkarni2023360fusionnerf}. While PERF \cite{wang2023perf} estimates depth using a 360-degree depth estimator, Omni-NeRF \cite{gu2022omni} and 360FusionNeRF \cite{kulkarni2023360fusionnerf} derive depth maps by projecting 2D image pixels onto a spherical surface and analyzing the intersections of rays with the scene geometry from multiple views. However, since our work involves images prior to their assembly into a 360-degree panorama and adheres to the pinhole camera model, research focused on spherical projections is not directly related to our scenario.

Compared to previous methods, our approach to computing architectural priors for indoor scenes and utilizing boundary loss shares similarities with Dense Depth Priors \cite{roessle2022dense} and NerfingMVS \cite{wei2021nerfingmvs}, as it follows the depth supervision approach using depth maps. Similar to StructNeRF \cite{chen2023structnerf}, we employ fundamental architectural principles to address non-textured areas. Unlike other studies, our approach imposes reliable geometric constraints in featureless regions of large architectural planes, without dependence on the inaccuracies associated with photometric consistency in these areas, thereby efficiently and effectively tackling challenges in low-feature regions.
Our depth regularization technique shares conceptual similarities with RegNeRF \cite{niemeyer2022regnerf} in its use of patches. However, our approach not only smooths out noise but also better preserves edges, enhancing depth consistency without sacrificing structural detail.

\section{\uppercase{Dataset}}
\label{sec:dataset}
\begin{figure*}[th!]
    \centering
    \captionsetup[subfigure]{labelformat=empty} 

    \begin{minipage}{0.1\textwidth}
        \centering
        \includegraphics[width=\linewidth]{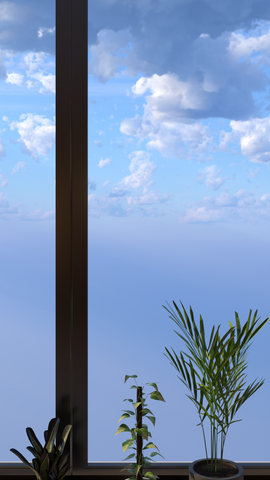}
    \end{minipage}%
    \begin{minipage}{0.1\textwidth}
        \centering
        \includegraphics[width=\linewidth]{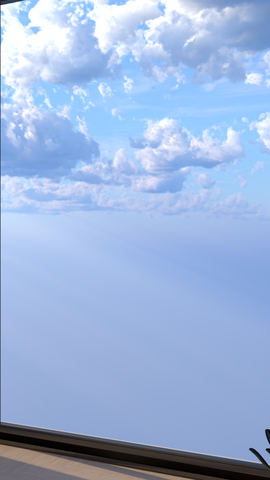}
    \end{minipage}%
    \begin{minipage}{0.1\textwidth}
        \centering
        \includegraphics[width=\linewidth]{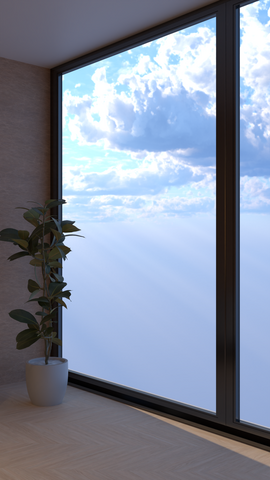}
    \end{minipage}%
    \begin{minipage}{0.1\textwidth}
        \centering
        \includegraphics[width=\linewidth]{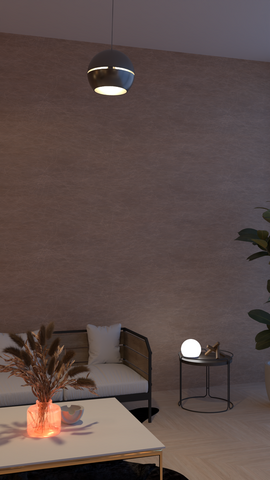}
    \end{minipage}%
    \begin{minipage}{0.1\textwidth}
        \centering
        \includegraphics[width=\linewidth]{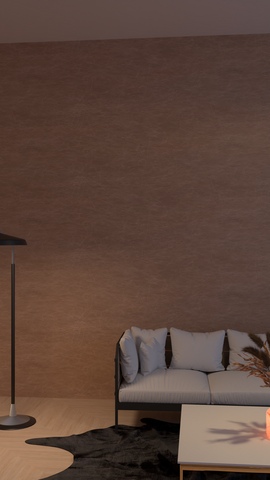}
    \end{minipage}%
    \begin{minipage}{0.1\textwidth}
        \centering
        \includegraphics[width=\linewidth]{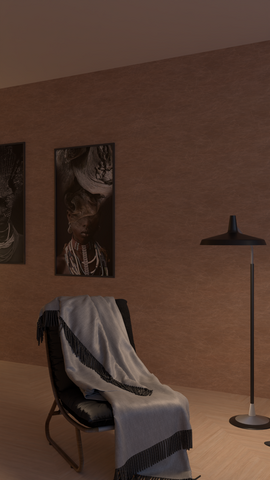}
    \end{minipage}%
    \begin{minipage}{0.1\textwidth}
        \centering
        \includegraphics[width=\linewidth]{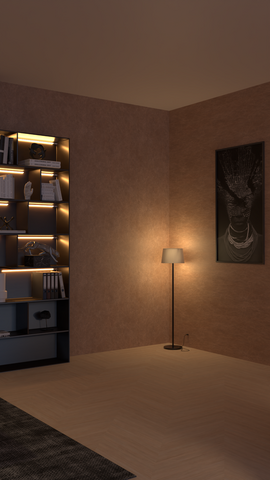}
    \end{minipage}%
    \begin{minipage}{0.1\textwidth}
        \centering
        \includegraphics[width=\linewidth]{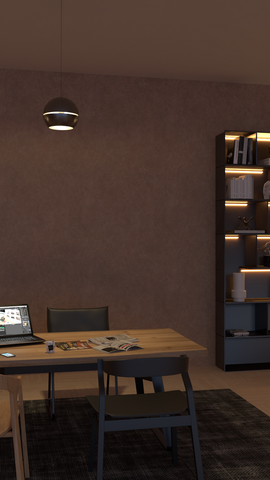}
    \end{minipage}%
    \begin{minipage}{0.1\textwidth}
        \centering
        \includegraphics[width=\linewidth]{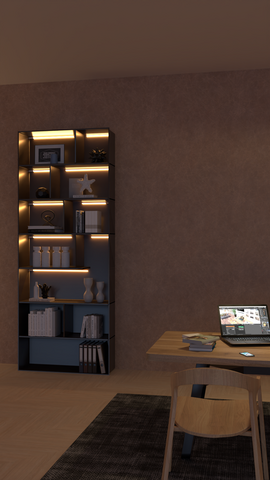}
    \end{minipage}%
    \begin{minipage}{0.1\textwidth}
        \centering
        \includegraphics[width=\linewidth]{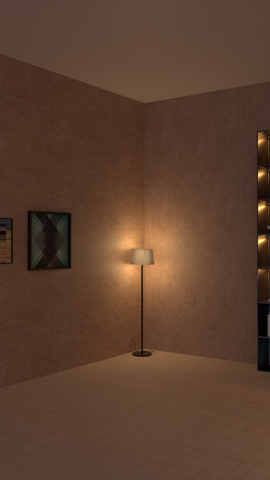}
    \end{minipage}%

    \begin{minipage}{0.1\textwidth}
        \centering
        \includegraphics[width=\linewidth]{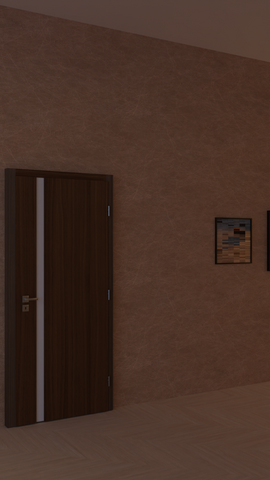}
    \end{minipage}%
    \begin{minipage}{0.1\textwidth}
        \centering
        \includegraphics[width=\linewidth]{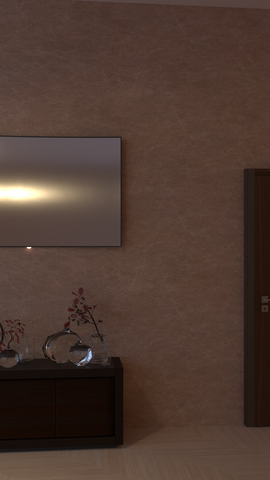}
    \end{minipage}%
    \begin{minipage}{0.1\textwidth}
        \centering
        \includegraphics[width=\linewidth]{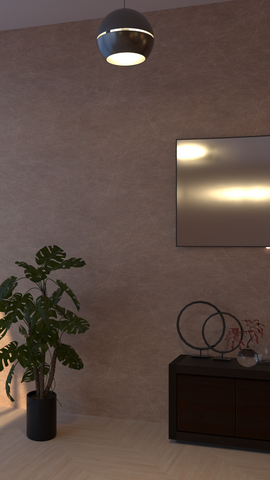}
    \end{minipage}%
    \begin{minipage}{0.1\textwidth}
        \centering
        \includegraphics[width=\linewidth]{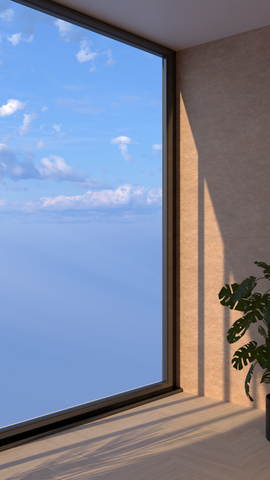}
    \end{minipage}%
    \begin{minipage}{0.1\textwidth}
        \centering
        \includegraphics[width=\linewidth]{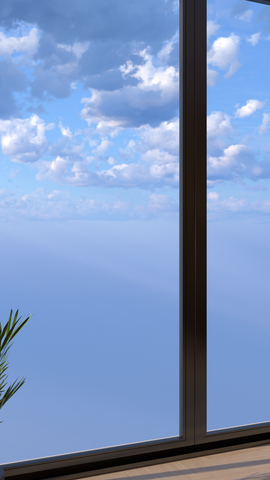}
    \end{minipage}%
    \begin{minipage}{0.1\textwidth}
        \centering
        \includegraphics[width=\linewidth]{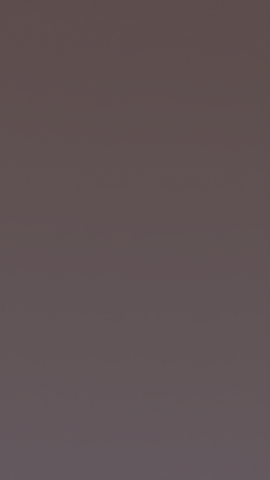}
    \end{minipage}%
    \begin{minipage}{0.1\textwidth}
        \centering
        \includegraphics[width=\linewidth]{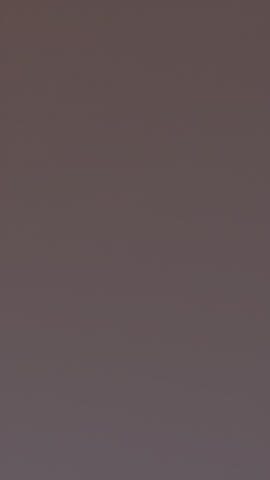}
    \end{minipage}%
    \begin{minipage}{0.1\textwidth}
        \centering
        \includegraphics[width=\linewidth]{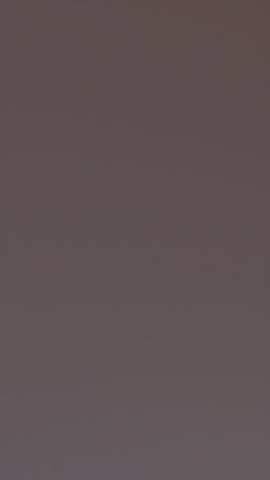}
    \end{minipage}%
    \begin{minipage}{0.1\textwidth}
        \centering
        \includegraphics[width=\linewidth]{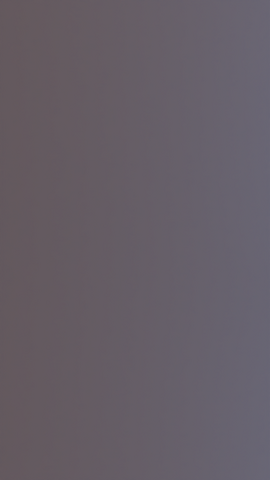}
    \end{minipage}%
    \begin{minipage}{0.1\textwidth}
        \centering
        \includegraphics[width=\linewidth]{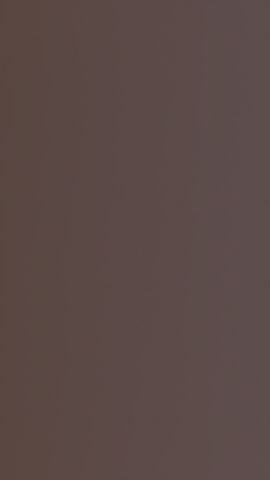}
    \end{minipage}%

    \begin{minipage}{0.1\textwidth}
        \centering
        \includegraphics[width=\linewidth]{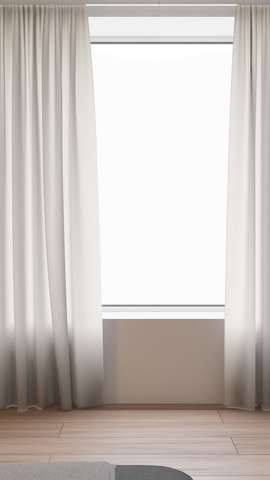}
    \end{minipage}%
    \begin{minipage}{0.1\textwidth}
        \centering
        \includegraphics[width=\linewidth]{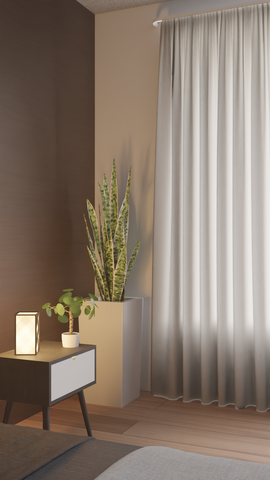}
    \end{minipage}%
    \begin{minipage}{0.1\textwidth}
        \centering
        \includegraphics[width=\linewidth]{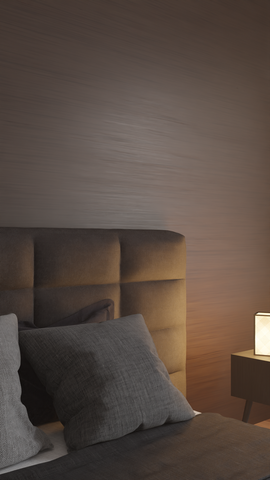}
    \end{minipage}%
    \begin{minipage}{0.1\textwidth}
        \centering
        \includegraphics[width=\linewidth]{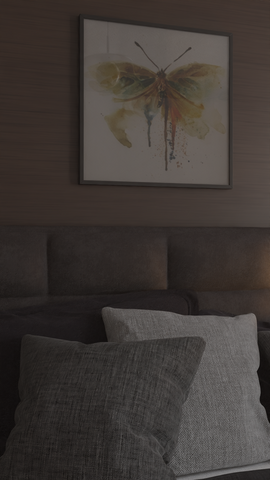}
    \end{minipage}%
    \begin{minipage}{0.1\textwidth}
        \centering
        \includegraphics[width=\linewidth]{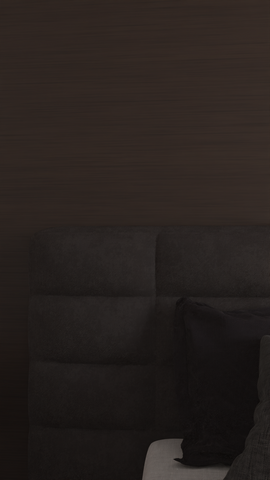}
    \end{minipage}%
    \begin{minipage}{0.1\textwidth}
        \centering
        \includegraphics[width=\linewidth]{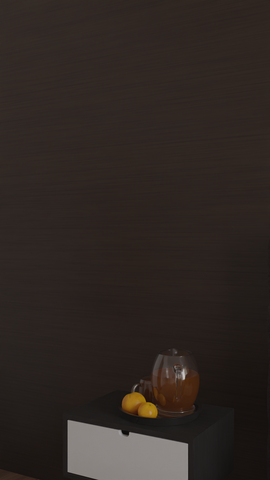}
    \end{minipage}%
    \begin{minipage}{0.1\textwidth}
        \centering
        \includegraphics[width=\linewidth]{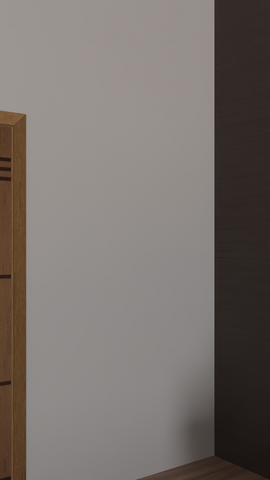}
    \end{minipage}%
    \begin{minipage}{0.1\textwidth}
        \centering
        \includegraphics[width=\linewidth]{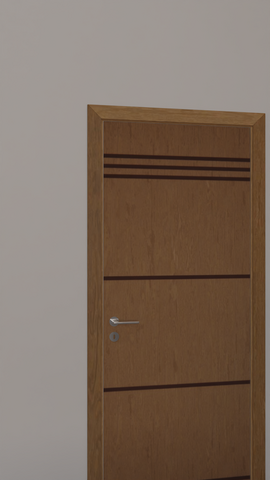}
    \end{minipage}%
    \begin{minipage}{0.1\textwidth}
        \centering
        \includegraphics[width=\linewidth]{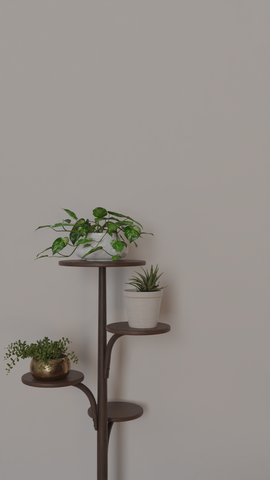}
    \end{minipage}%
    \begin{minipage}{0.1\textwidth}
        \centering
        \includegraphics[width=\linewidth]{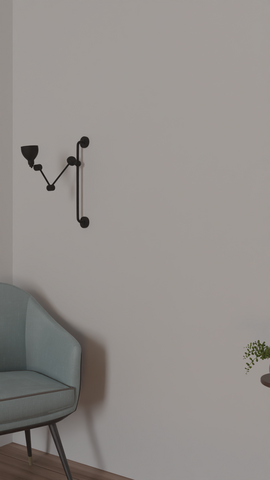}
    \end{minipage}%

    \begin{minipage}{0.1\textwidth}
        \centering
        \includegraphics[width=\linewidth]{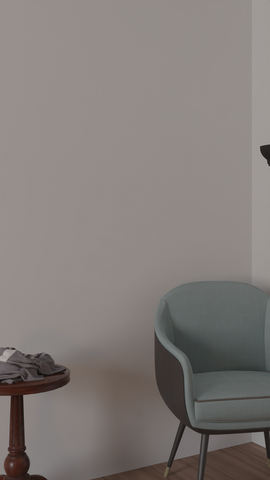}
    \end{minipage}%
    \begin{minipage}{0.1\textwidth}
        \centering
        \includegraphics[width=\linewidth]{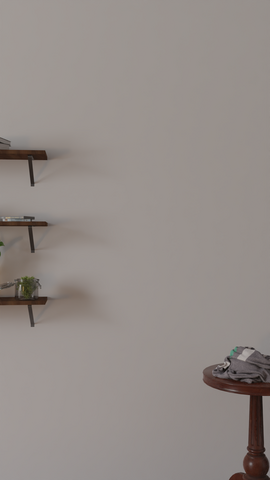}
    \end{minipage}%
    \begin{minipage}{0.1\textwidth}
        \centering
        \includegraphics[width=\linewidth]{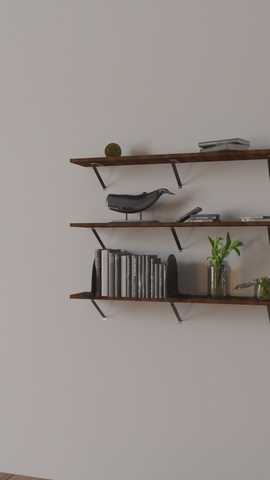}
    \end{minipage}%
    \begin{minipage}{0.1\textwidth}
        \centering
        \includegraphics[width=\linewidth]{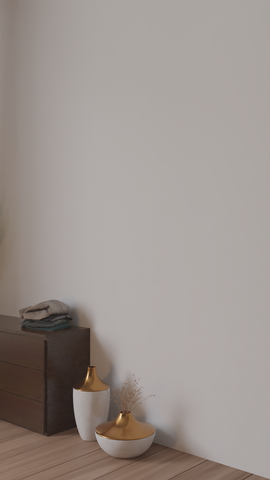}
    \end{minipage}%
    \begin{minipage}{0.1\textwidth}
        \centering
        \includegraphics[width=\linewidth]{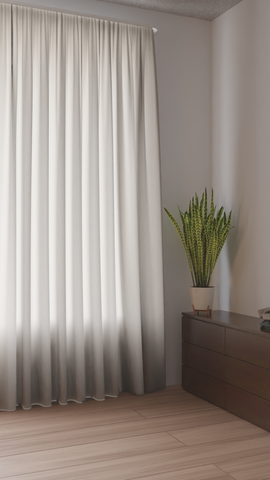}
    \end{minipage}%
    \begin{minipage}{0.1\textwidth}
        \centering
        \includegraphics[width=\linewidth]{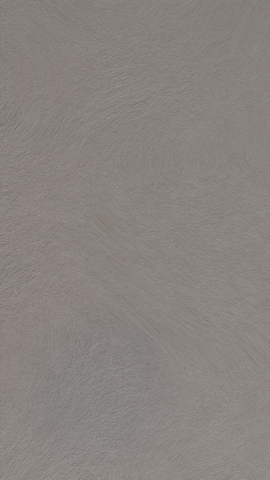}
    \end{minipage}%
    \begin{minipage}{0.1\textwidth}
        \centering
        \includegraphics[width=\linewidth]{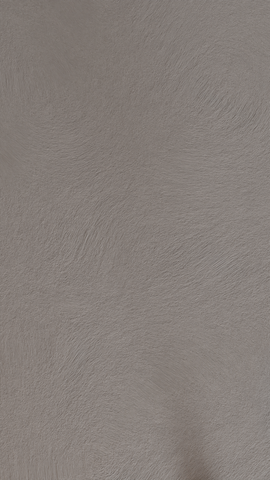}
    \end{minipage}%
    \begin{minipage}{0.1\textwidth}
        \centering
        \includegraphics[width=\linewidth]{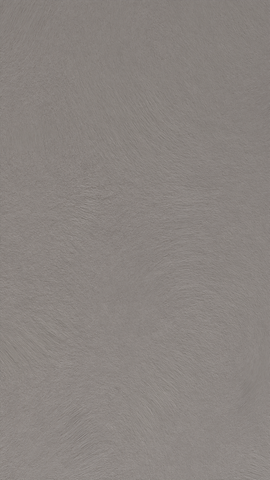}
    \end{minipage}%
    \begin{minipage}{0.1\textwidth}
        \centering
        \includegraphics[width=\linewidth]{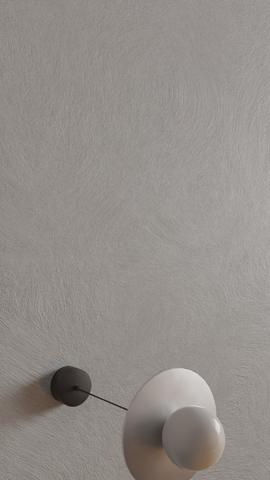}
    \end{minipage}%
    \begin{minipage}{0.1\textwidth}
        \centering
        \includegraphics[width=\linewidth]{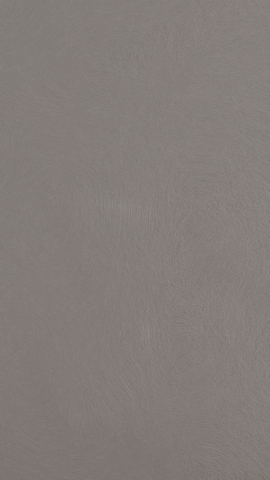}
    \end{minipage}%
    
    \caption{Raw images captured with a pinhole camera model, showing unstitched frames prior to assembly into a 360-degree panorama. The living room is depicted in the first two rows and the bedroom in the last two. The first 15 images (from top left to bottom right) depict a 360-degree horizontal sweep, while the final 5 images capture the upper surroundings.}
    \label{fig:unstiched_panorama_livingroom}
\end{figure*}

Our focus is on capturing indoor scenarios using a 360-degree camera. To cover the entire space, we recommend a mounted, movable setup. Rather than working with a stitched 360-degree panorama, we propose using a series of individual raw views prior to their assembly (see Figure \ref{fig:unstiched_panorama_livingroom}). While unconventional, this approach has the potential to significantly improve the accuracy of extrinsic and intrinsic data compared to a stitched panorama—essential for the NeRF pipeline—and, consequently, enhance the overall quality of NeRF-rendered scenes \cite{gu2022omni}.

Hence, we generated a custom dataset in Blender comprising two synthetic indoor scenes: a bedroom (6×8×3.8 m) and a living room (10×10×3.4 m). Both scenes are modeled with the floor at Z = 0 and the origin at the center of the floor, with orthogonal coordinate axes and the positive Z-axis extending upward. Individual images of an unstitched 360-degree panorama were captured using Blender's perspective camera with a $27^\circ$ horizontal and $40^\circ$ vertical field of view. Each 360-degree horizontal sweep consisted of 15 images, spaced at $24^\circ$ intervals, with 5 additional images covering the ceiling by first rotating the camera upward and tilting it in four directions. A $3^\circ$ overlap between adjacent images ensured seamless assembly. Cameras were positioned in a grid pattern across the scene with random noise added for realism.

The living room dataset comprises 1200 training images and 540 evaluation images, while the bedroom dataset includes 840 training images and 300 evaluation images. Each RGB image, at a resolution of 1080×1920 pixels, is provided with camera parameters, depth maps, and segmentation maps for planar architectural surfaces such as floors, ceilings, and walls.

\section{\uppercase{Methodology}}
\label{sec:methodology}
In this section, we outline our methodology for enhancing NeRF rendering quality in indoor environments, specifically focusing on reducing cloudy artifacts, commonly called ''floaters'', that often appear on featureless surfaces. Our approach incorporates custom depth estimation techniques for planar architectural surfaces, such as walls, floors, and ceilings, along with a loss function tailored for boundary regions. Moreover, we propose a depth regularization technique that complements the previous approach by refining rendering quality across the entire scene.

We begin by discussing depth supervision techniques, followed by an introduction of a novel depth estimation method explicitly designed for planar architectural surfaces in indoor scenes. Next, we introduce a boundary loss function that enforces spatial constraints, improving depth accuracy along architectural boundaries. Finally, we outline our custom patch-based depth regularization method.

\subsection{Depth Supervision}
\label{sec:depth_supervision} 
Depth supervision is an effective approach to mitigate floating artifacts by comparing rendered and ground truth depth \cite{wang2023digging}. It constrains the density distribution, enforcing geometric consistency.

Specifically, the color $\hat{C}(\textbf{r})$ and depth $\hat{D}(\textbf{r})$ of a pixel along a ray $\mathbf{r}$ are rendered by NeRFs as follows:

\begin{equation}
\hat{C}(\textbf{r}) = \sum_{i=1}^{N} w_i \mathbf{c}_i,
\label{eqn:rendered_color}
\end{equation}

\begin{equation}
\hat{D}(\textbf{r}) = \sum_{i=1}^{N} w_i t_i,
\label{eqn:rendred_depth}
\end{equation}

where $\hat{C}(\textbf{r})$ is the final color rendered for the pixel along ray $\mathbf{r}$, and $\hat{D}(\textbf{r})$ is the estimated depth from the camera to the pixel along ray $\textbf{r}$. Here, $N$ denotes the number of samples along $\mathbf{r}$.

The weight for the $i$-th sample, representing the contribution of a sample $i$ along the ray $\textbf{r}$ to the final color and depth values for the corresponding pixel, is defined as:

\begin{equation} w_i = T_i \alpha_i.
\label{eqn:weight}
\end{equation}

The transmittance $T_i$ at sample $i$, indicating the probability of light reaching the sample unimpeded, is defined as:

\begin{equation}
T_i = \exp \left( - \sum_{j=1}^{i-1} \sigma_j \Delta_j \right).
\label{eqn:transmittance}
\end{equation}

The opacity $\alpha_i$ at sample $i$ represents the likelihood that light is absorbed or scattered at sample $i$ and is given by:

\begin{equation}
\alpha_i = 1 - \exp(- \sigma_i \Delta_i).
\label{eqn:opcacity}
\end{equation}

Further, $\sigma_i$ is the volume density at sample $i$ and $\Delta_i = t_{i+1} - t_i$ is the distance between adjacent samples. Here, $\mathbf{c}_i$ represents the RGB color, and $t_i$ is the distance from the camera origin to the $i$-th sample.

NeRFs are optimized by enforcing rendered color consistency through a photometric loss function, commonly defined as the Mean Squared Error (MSE) between the rendered and ground truth pixel colors \cite{rabby2023beyondpixels}:

\begin{equation}
\mathcal{L}_{\text{color}} = \sum_{\textbf{r} \in \mathcal{R}} \left| \hat{C}(\textbf{r}) - C(\textbf{r}) \right|^2_2,
\label{eqn:color_loss}
\end{equation}

where $\mathcal{R}$ represents the set of rays in each training batch, and $C(\textbf{r})$ and $\hat{C}(\textbf{r})$ denote the ground truth and predicted RGB colors for each ray $\textbf{r}$, respectively.

Depth supervision is applied by combining this photometric loss with an additional depth loss:

\begin{equation}
\mathcal{L} = \lambda_{\text{color}} \mathcal{L}_{\text{color}} + \lambda_{\text{depth}} \mathcal{L}_\text{depth},
\label{eqn:depth_superv}
\end{equation}

where $\lambda_{\text{color}}$ and $\lambda_{\text{depth}}$ are weighting factors that balance the contributions of the photometric and depth losses, respectively.

In this work, we utilize an MSE loss to compare the rendered and ground truth depths:

\begin{equation}
\mathcal{L}_\text{depth} = \sum_{\textbf{r}\in \mathcal{R}} \left| \hat{D}(\textbf{r}) - D(\textbf{r}) \right|^2_2.
\label{eqn:depth_loss_mse}
\end{equation}

Here, $D(\mathbf{r})$ and $\hat{D}(\mathbf{r})$ are the ground truth and predicted depths, respectively, for ray $\mathbf{r}$ from the ray batch $\mathcal{R}$.

However, depth supervision relies on accurate ground truth depth data, which is often difficult to obtain in real-world scenarios \cite{ming2021deep}. A common approach for acquiring depth priors for NeRF is through Structure from Motion techniques, particularly COLMAP, which generates depth information as a byproduct of camera pose estimation \cite{roessle2022dense}. Since Structure from Motion methods rely on keypoint matching across multiple images to establish correspondences, they often struggle on textureless areas lacking distinctive visual features—a challenge especially pronounced in indoor environments dominated by uniform architectural surfaces.

\subsection{Depth Estimation for Planar Architectural Surfaces}
\label{sec:depth_planar-surfaces}
We propose a fast, simple, and computationally efficient method to estimate depth in featureless indoor regions such as walls, floors, and ceilings. The approach assumes the Z-axis origin is calibrated to lie on the floor plane. If not, three non-collinear camera positions at a constant height (e.g., tripod-mounted) must be available. Room height must also be known, along with semantic segmentation for wall, floor, and ceiling classes, which can be efficiently generated using pretrained models such as DeepLab \cite{deeplabv3plus2018}.

Depth computation leverages the NeRF ray representation \cite{mildenhall2021nerf}, defined as:
\begin{equation}
\mathbf{r}(t) = \mathbf{o} + t\mathbf{d}_{\text{unit}},
\label{eqn}
\end{equation}
where $\mathbf{o}$ is the camera origin and $\mathbf{d}_{\text{unit}}$ is a unit vector representing the ray direction. Using camera parameters, the Euclidean depth $t$ of a pixel $\mathbf{P}$ is determined by setting one known component of its 3D world coordinate (e.g., the Z-coordinate $P_z$, which represents height in 3D space). Knowing $t$ allows the recovery of the remaining 3D coordinates of $\mathbf{P}$.

For floors and ceilings, if the Z-axis origin lies on the floor, floor depth can be computed directly by setting $P_z = 0$, while ceiling depth is computed by setting $P_z$ to the ceiling height. Without calibration, the plane equation $s_{\text{cam}}$ is derived from three non-collinear camera positions. Parallel planes are then calculated at distances equal to the camera height above and below $s_{\text{cam}}$. The floor plane $s_{\text{floor}}$ is identified as the parallel plane that intersects the ray corresponding to an arbitrary floor pixel. Next, the ceiling plane $s_{\text{ceil}}$ is determined similarly, accounting for the ceiling height relative to the camera.

To estimate wall depths, border pixels where walls meet the ceiling and floor are first identified. Three non-collinear points (two from one border and one from the other) are selected to define the wall plane $s_{\text{wall}}$. Finally, depth for walls, ceiling, and floor is computed as the Euclidean distance from the ray origins of pixels belonging to the corresponding segmentation classes to their intersection points with the respective planes.

\subsection{Boundary Loss For Architectural Surfaces} 
\label{sec:boundary_loss}

\begin{figure}[h]
    \includegraphics[width=0.5\textwidth]{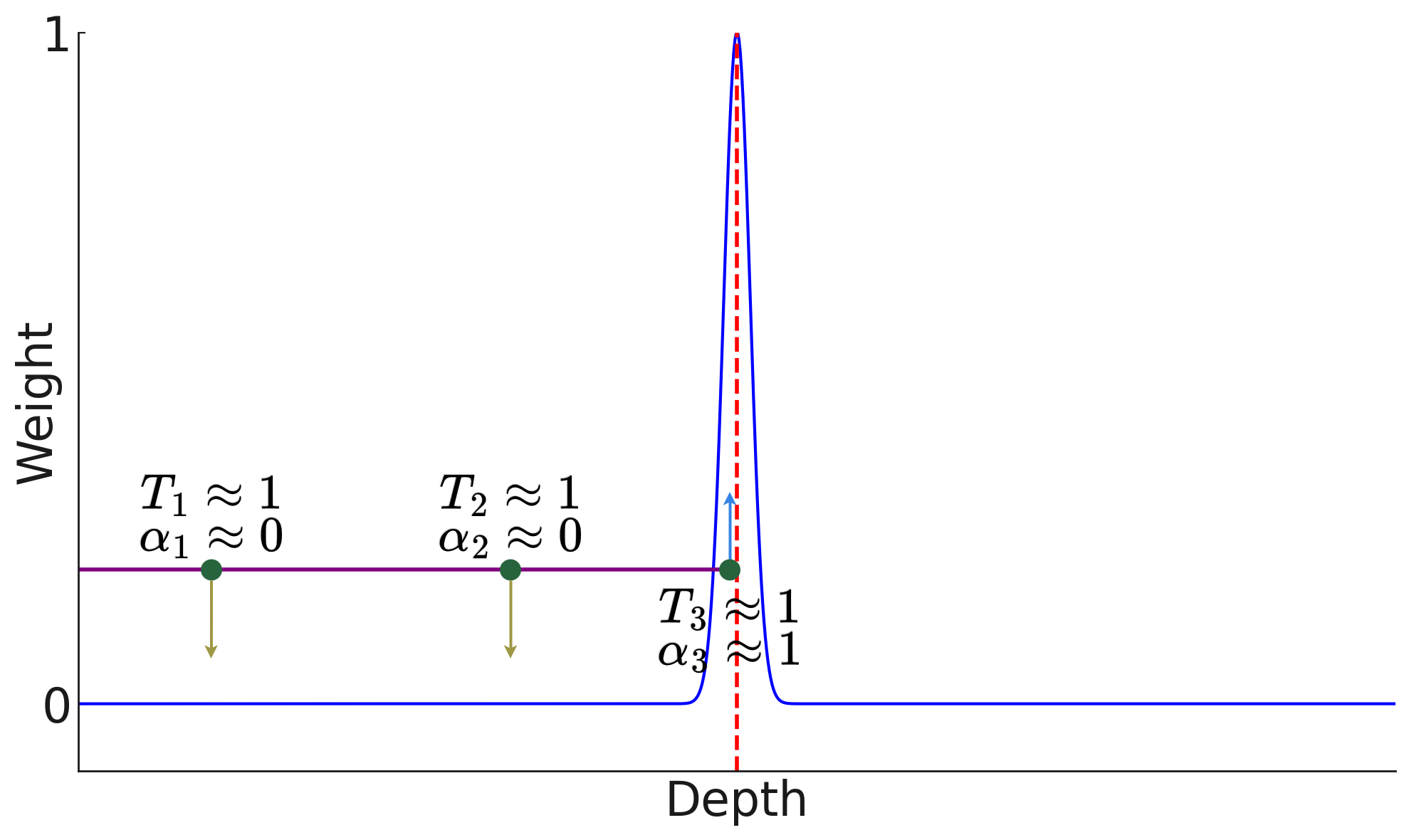}
    \caption{Illustration of a Gaussian distribution modeling the weight \( w_i \) along a ray which hits the boundary surface (e.g., a wall) depicted by the red dotted line. The purple solid line indicates the ray with the green dots representing samples.}
    \label{fig:gaussian_distribution_ray}
\end{figure}

When a ray travels through open space within a room and does not intersect any surface, its transmittance $T_i$ remains high, meaning the ray continues unimpeded through the scene, while its opacity $\alpha_i$ remains low, reflecting the absence of intersecting material. This combination of high transmittance and low opacity produces low weights along the ray’s path (see eq.\ (\ref{eqn:weight})), as there is minimal interaction to indicate boundaries, as depicted by the yellow downward arrows in Figure \ref{fig:gaussian_distribution_ray}.

However, as the ray reaches a boundary surface (like a wall or ceiling), the interaction characteristics change. The transmittance $T_i$ remains high initially, as the ray is still progressing through space, but the opacity $\alpha_i$ begins to rise due to the increasing material density encountered at the boundary. As illustrated by the upward blue arrow in Figure \ref{fig:gaussian_distribution_ray}, this increase in opacity correlates with higher weights near the boundary, highlighting the role of these architectural surfaces in defining the spatial limits within the scene. When the ray finally intersects a boundary surface, the weights along the ray peak, often reaching a maximum (e.g., a weight of 1), as the ray’s traversal is effectively complete \cite{szeliski2022computer}.

Based on these observations, we introduce a boundary loss function that leverages our architectural depth priors:

\begin{equation}
    \mathcal{L}_{bound} = \sum_{\textbf{r} \in \mathcal{R}} \sum_{i} \left( w_i - e^{\left( - \dfrac{(t_i - D(\textbf{r}))^2}{2{\sigma}^2} \right)} \right)^2,
    \label{eqn:bound_loss}
\end{equation}
where $D(\textbf{r})$ is the ground truth depth of ray $\textbf{r} $ from ray batch $\mathcal{R}$, $w_i $ is a weight corresponding to a point, which is sampled on the ray $\textbf{r}$ at the distance $t_i$ from the ray origin.

For pixels corresponding to architectural surfaces, the boundary loss penalizes weights of samples far from the surface and boosts weights close to it, enforcing correct architectural constraints.

\subsection{Patch-Based Depth Regularization}
\label{sec:depth_regularization}

To complement our depth supervision on planar architectural surfaces and mitigate rendering irregularities beyond these regions, we draw inspiration from Reg-NeRF \cite{niemeyer2022regnerf} and propose a depth regularization method that operates on image patches. This approach promotes smooth and consistent depth predictions across rendered views, effectively reducing noise and artifacts while preserving essential structural details. Specifically, we apply a bilateral \cite{tomasi1998bilateral} or joint bilateral filter \cite{he2012guided} to regularize the depth within each patch.

\textbf{Filtering the Depth Patch}. We begin by applying a bilateral or joint bilateral filter to a rendered depth patch $\hat{D}(p)$, where $p$ is a patch from the set $\mathbf{P}$. The bilateral filter accounts for both spatial proximity and depth similarity, while the joint bilateral filter additionally considers intensity similarity in the corresponding RGB image. This method ensures that the smoothing of depth values respects the structural edges present in the image.

\textbf{Computing the Regularization Loss}. For each depth patch $p \in \mathbf{P}$, we calculate the Mean Squared Error (MSE) between the original rendered depth patch $\hat{D}(p)$ and the filtered depth patch $\mathcal{F}(\hat{D}(p))$. We then compute the average of these MSE losses across all patches in $P$ to obtain a single regularization term:

\begin{equation}
\mathcal{L}_{\text{reg}} = \frac{1}{|\mathbf{P}|} \sum_{p \in \mathbf{P}} \frac{1}{|p|} \sum_{i,j} \left( \hat{D}(p_{ij}) - \mathcal{F}(\hat{D}(p))_{ij} \right)^2.
\end{equation}

This regularization term is incorporated into the total loss function in the same manner as depth supervision (see eq. \ref{eqn:depth_superv}).

\section{\uppercase{Experiments}}
\label{sec:results}

To evaluate the effectiveness of our approach in enhancing NeRF rendering quality in indoor environments, we conducted a series of experiments using Instant-NGP \cite{muller2022instant} within the Nerfstudio framework \cite{tancik2023nerfstudio}. Instant-NGP was chosen for its hash encoding, which captures objects of varying sizes, and occupancy grids, which focus computation on meaningful areas in indoor scenes with significant empty space. Given computational constraints, we downsampled our datasets by a factor of two, resulting in a final resolution of 540x960 pixels. All models were trained on an NVIDIA GeForce RTX 3090 GPU using the Adam optimizer with parameters $\beta_1 = 0.9$, $\beta_2 = 0.99$, and $\epsilon = 10^{-8}$. Neither weight decay nor gradient clipping was applied. We set the hash table size $T$ to 22, the maximum
resolution $N_{\max}$ to 32,768, the density MLP to a depth of 2 and a width of 64, and the color MLP to a depth of 2 and a width of 128. These values were determined through hyperparameter optimization. All other hyperparameters followed Nerfstudio defaults. It took 200,000 iterations to train the models with photometric loss on both scenes, with performance plateauing beyond this point.

\textbf{Patch-Based Depth Regularization}. We implemented patch-based regularization using the open-source library Kornia, utilizing its default parameters for bilateral and joint bilateral filters: a $9 \times 9$ kernel size, range sigma ($\sigma_{\text{color}}$) of 10 to control intensity similarity, and spatial sigma ($\sigma_{\text{space}}$) of $75 \times 75$ to define the spatial extent of the filter. 

Training with patches significantly extended the process, requiring additional time for the network to capture global image structure. To address this, the model was first trained to convergence with photometric loss, followed by patch-based regularization to refine details. We trained with a patch size of 16, as larger patches (32 and 64) remained undertrained even after 400,000 iterations and significantly increased training time. The best results were achieved with $\lambda_{\text{color}} = 1$ and $\lambda_{\text{reg}} = 10^{-7}$ for both bilateral and joint bilateral loss. Models using these parameters converged in 280,000 iterations. For comparison, we implemented patch similarity constraints as described in RegNeRF \cite{niemeyer2022regnerf}, following the same training strategy.

\textbf{Depth Supervision with Planar Architectural Depth Priors}. As a preprocessing step, we computed depth estimates for the floor, ceiling, and walls using semantic segmentation generated in Blender. To evaluate the accuracy of these depth priors, we compared them against Blender-generated depth maps as ground truth. The results demonstrated high accuracy, with Root Mean Square Error (RMSE) values of 2.786 mm for the bedroom scene and 3.201 mm for the living room scene.

Next, we incorporated these architectural depth priors into the training process. Instant-NGP was trained on both scenes with depth supervision, employing MSE and BoundL loss alongside the priors, continuing each model until convergence. For models utilizing BoundL, we set \( \delta \) to 1 mm. The weight values for the losses were set to \( \lambda_{\text{color}} = 10 \) and \( \lambda_{\text{depth}} = 10 \) for the bedroom scene, and \( \lambda_{\text{color}} = 1 \) and \( \lambda_{\text{depth}} = 1 \) for the living room scene. Depth-supervised models converged in only 120,000 iterations, demonstrating the efficiency of incorporating planar architectural depth priors into the training process.

\section{\uppercase{Results}}
\begin{figure}[h!]
    \centering
    \captionsetup[subfigure]{labelformat=empty}
    \begin{minipage}{0.16\textwidth}
        \centering
        \includegraphics[width=\linewidth]{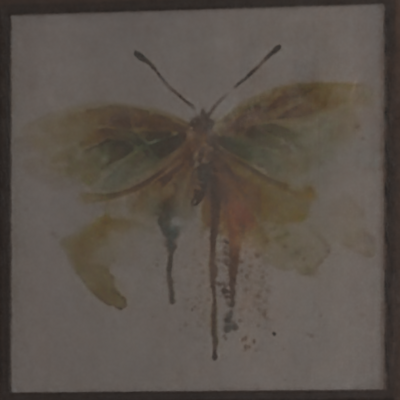}
    \end{minipage}%
    \begin{minipage}{0.16\textwidth}
        \centering
        \includegraphics[width=\linewidth]{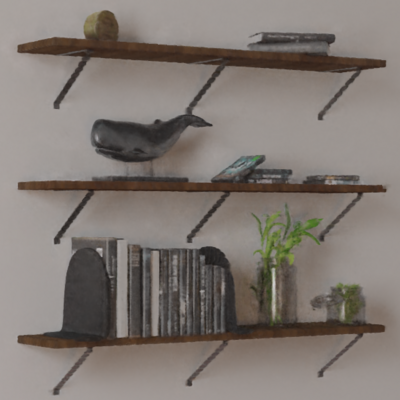}
    \end{minipage}%
    \begin{minipage}{0.16\textwidth}
        \centering
        \includegraphics[width=\linewidth]{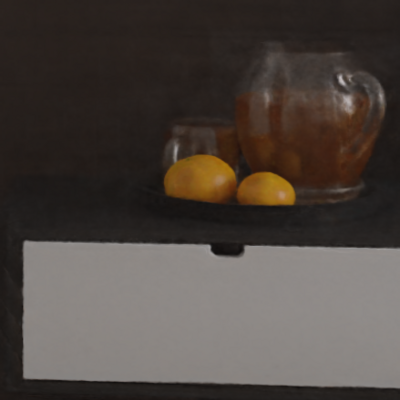}
    \end{minipage}%
    
    \begin{minipage}{0.16\textwidth}
        \centering
        \includegraphics[width=\linewidth]{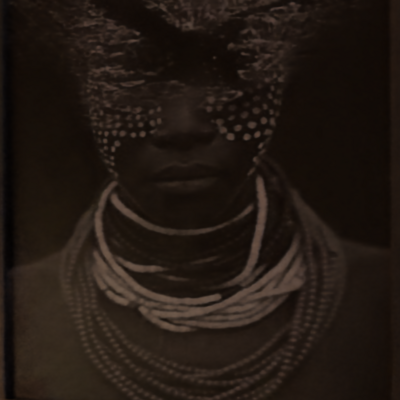}
    \end{minipage}%
    \begin{minipage}{0.16\textwidth}
        \centering
        \includegraphics[width=\linewidth]{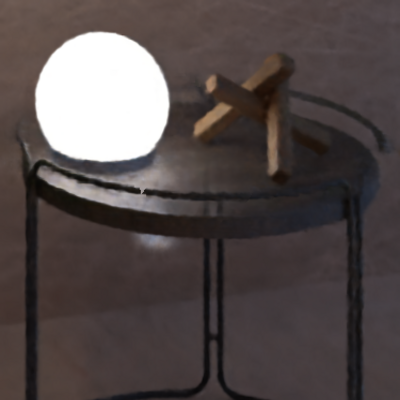}
    \end{minipage}%
    \begin{minipage}{0.16\textwidth}
        \centering
        \includegraphics[width=\linewidth]{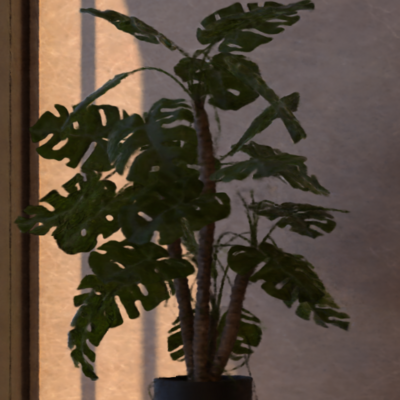}
    \end{minipage}%
    \caption{Renderings with Instant-NGP trained on our 360-degree indoor dataset using photometric loss show high visual fidelity on detail-rich areas.}
    \label{fig:details}
\end{figure}

\begin{figure}[h!]
    \centering
    \captionsetup[subfigure]{labelformat=empty}
    \begin{minipage}{0.12\textwidth}
        \centering
        \includegraphics[width=\linewidth]{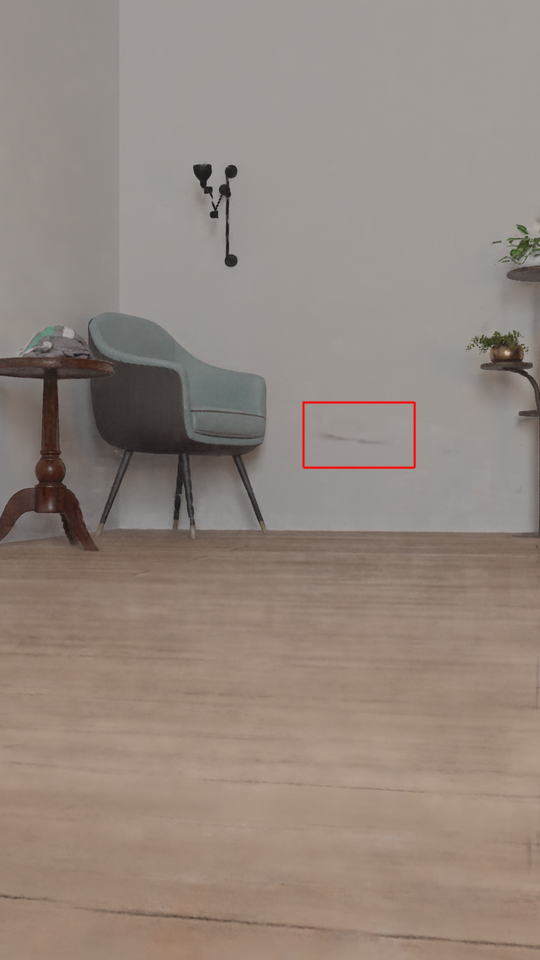}
    \end{minipage}%
    \begin{minipage}{0.12\textwidth}
        \centering
        \includegraphics[width=\linewidth]{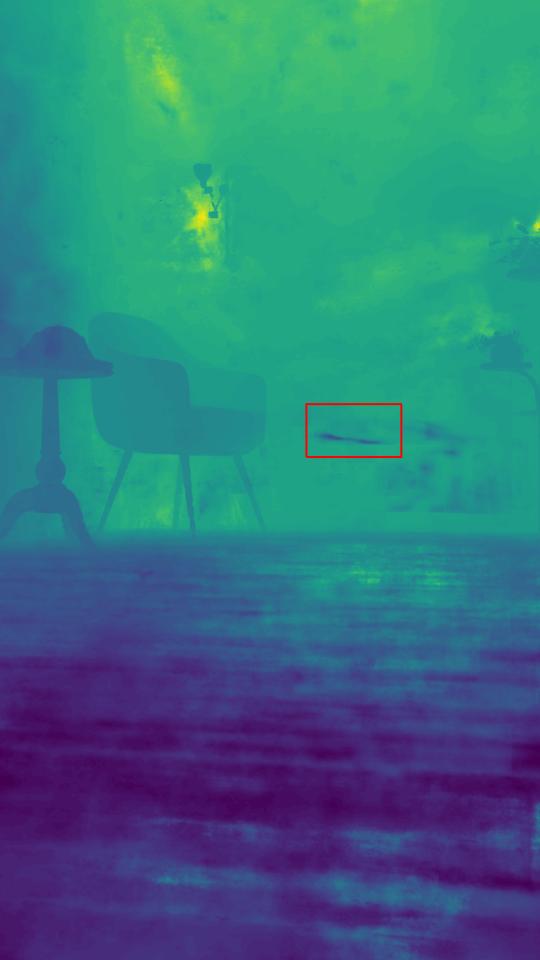}
    \end{minipage}%
    \begin{minipage}{0.12\textwidth}
        \centering
        \includegraphics[width=\linewidth]{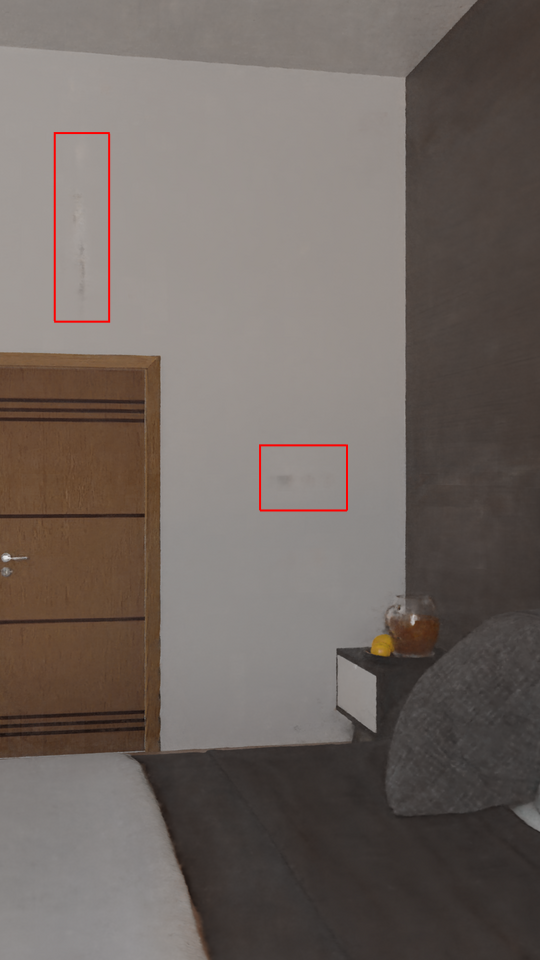}
    \end{minipage}%
    \begin{minipage}{0.12\textwidth}
        \centering
        \includegraphics[width=\linewidth]{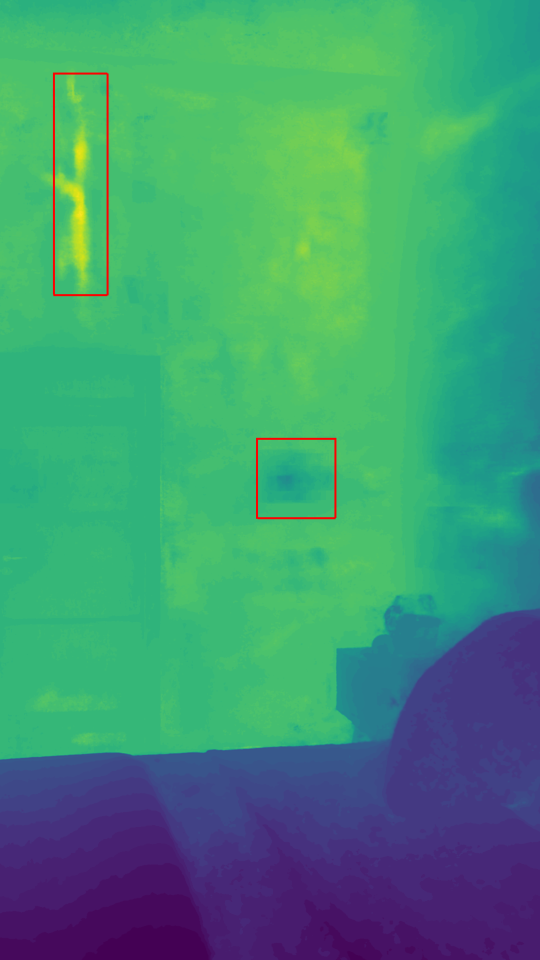}
    \end{minipage}%

    \begin{minipage}{0.12\textwidth}
        \centering
        \includegraphics[width=\linewidth]{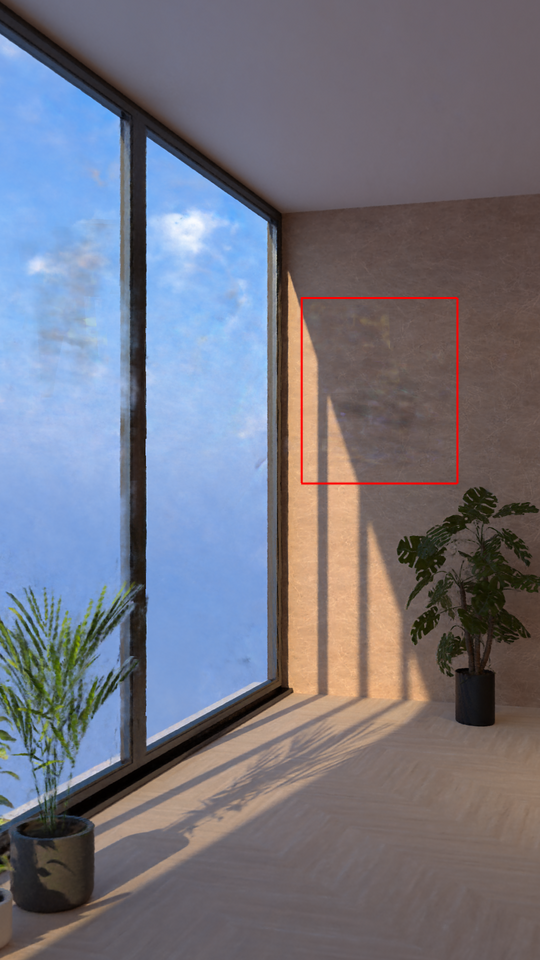}
    \end{minipage}%
    \begin{minipage}{0.12\textwidth}
        \centering
        \includegraphics[width=\linewidth]{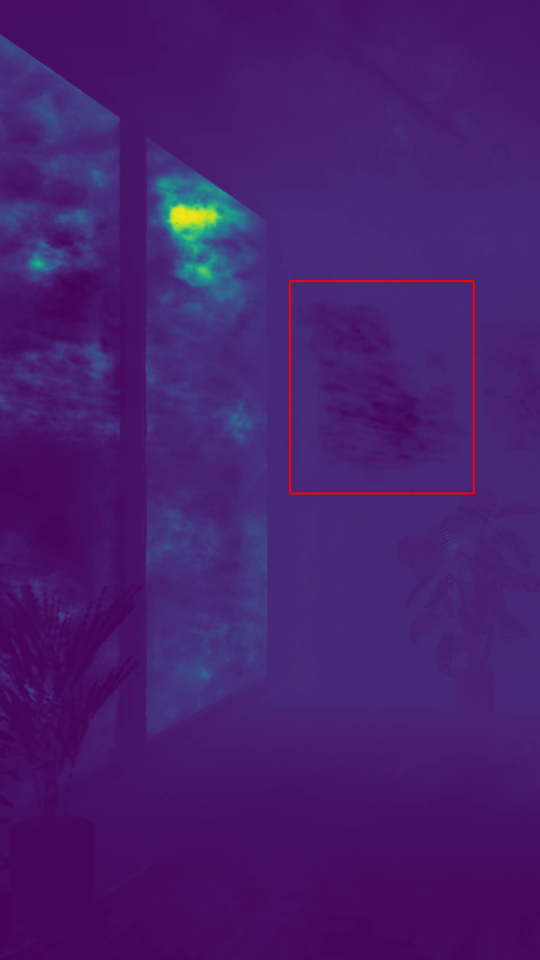}
    \end{minipage}%
    \begin{minipage}{0.12\textwidth}
        \centering
        \includegraphics[width=\linewidth]{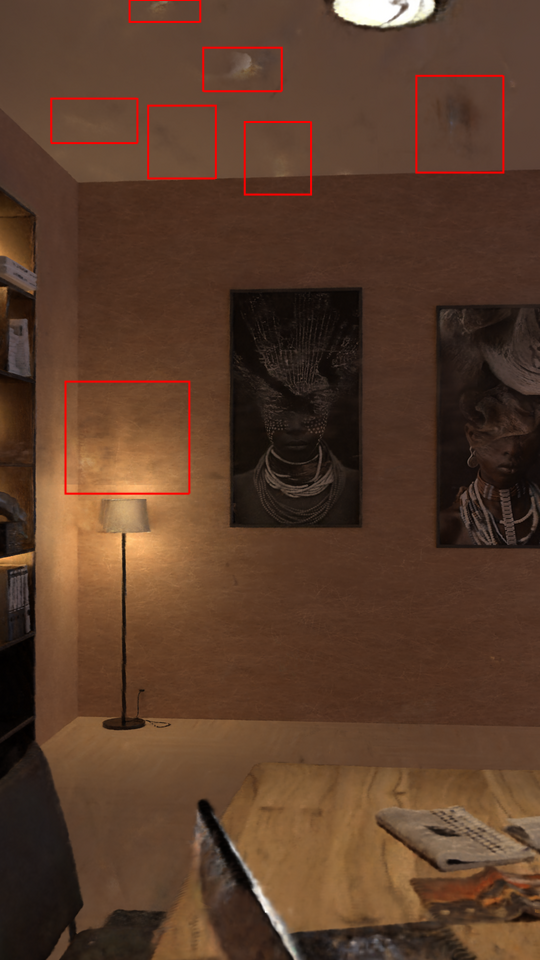}
    \end{minipage}%
    \begin{minipage}{0.12\textwidth}
        \centering
        \includegraphics[width=\linewidth]{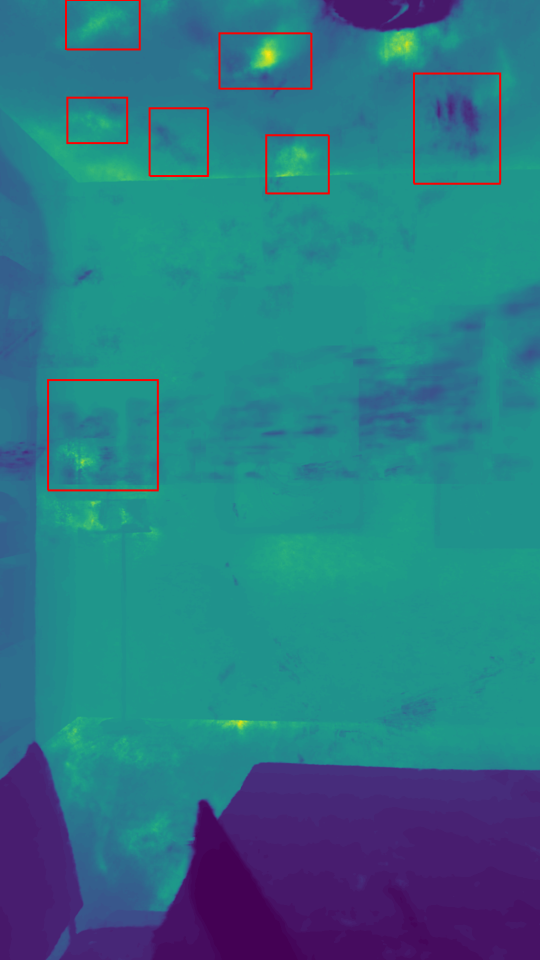}
    \end{minipage}%
    \caption{Renderings produced by Instant-NGP trained on our 360-degree indoor dataset with photometric loss are displayed alongside their corresponding depth maps. Red bounding boxes highlight floaters in front of walls, ceilings, or floors, caused by incorrect depth estimations.}
    \label{fig:floaters}
\end{figure}

Renderings produced by our baseline model, which relies solely on photometric loss, confirm our initial observation: ''floaters'' are more common on textureless surfaces like walls, floors, and ceilings (see Figure \ref{fig:floaters}). In contrast, objects with rich visual features—such as plants, books, and paintings—exhibit fewer floaters, as shown in Figure \ref{fig:details}. Notably, cloudy artifacts consistently align with incorrect depth estimations. This outcome underscores the limitations of NeRFs when relying solely on RGB optimization signals to accurately predict geometric constraints in featureless regions. Interestingly, some inconsistencies in the rendered depth maps did not produce visible artifacts in the color image, indicating a degree of tolerance in NeRF’s volume rendering.

\begin{figure}[h!]
    \centering
    \captionsetup[subfigure]{labelformat=parens}
    \begin{minipage}{0.12\textwidth}
        \centering
        \includegraphics[width=\linewidth]{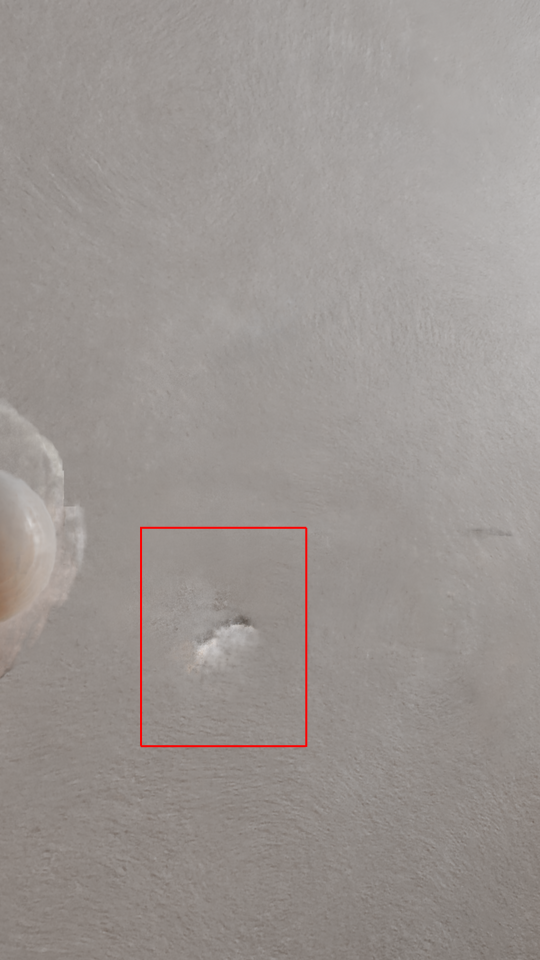}
        \subcaption{RGB}
    \end{minipage}%
        \begin{minipage}{0.12\textwidth}
        \centering
        \includegraphics[width=\linewidth]{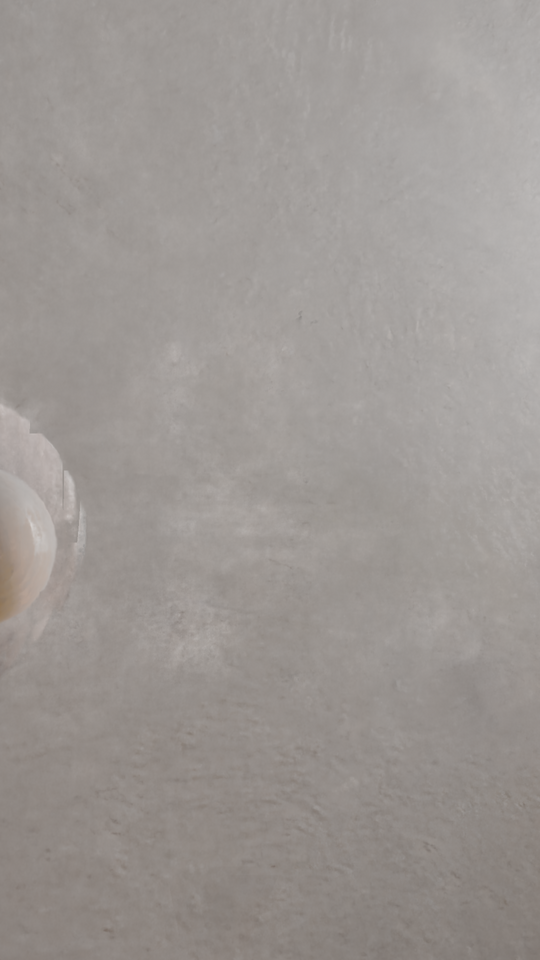}
        \subcaption{BoundL}
    \end{minipage}%
        \begin{minipage}{0.12\textwidth}
        \centering
        \includegraphics[width=\linewidth]{images/evaluation/bedroom/bad/c_00167.png}
        \subcaption{RGB}
    \end{minipage}%
    \begin{minipage}{0.12\textwidth}
        \centering
        \includegraphics[width=\linewidth]{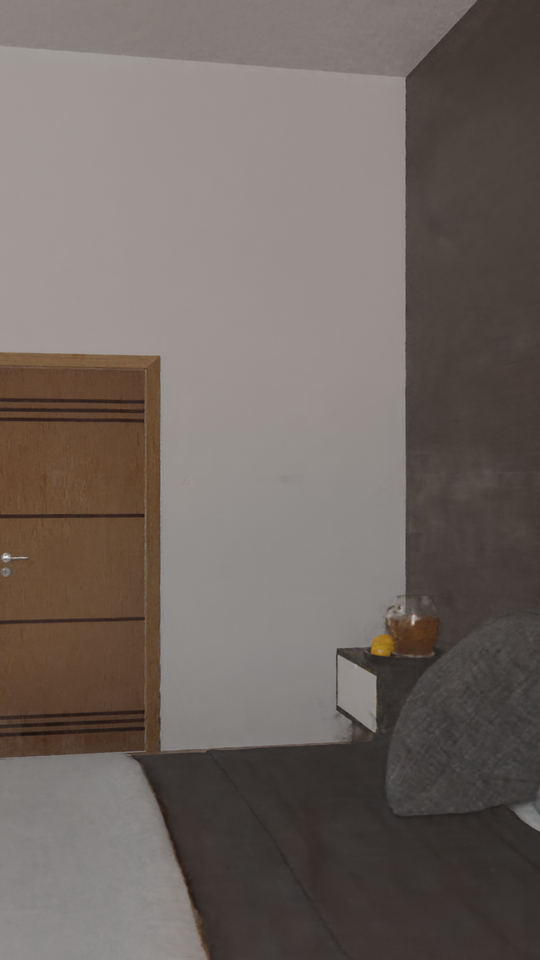}
        \subcaption{BoundL}
    \end{minipage}%
    
    \begin{minipage}{0.12\textwidth}
        \centering
        \includegraphics[width=\linewidth]{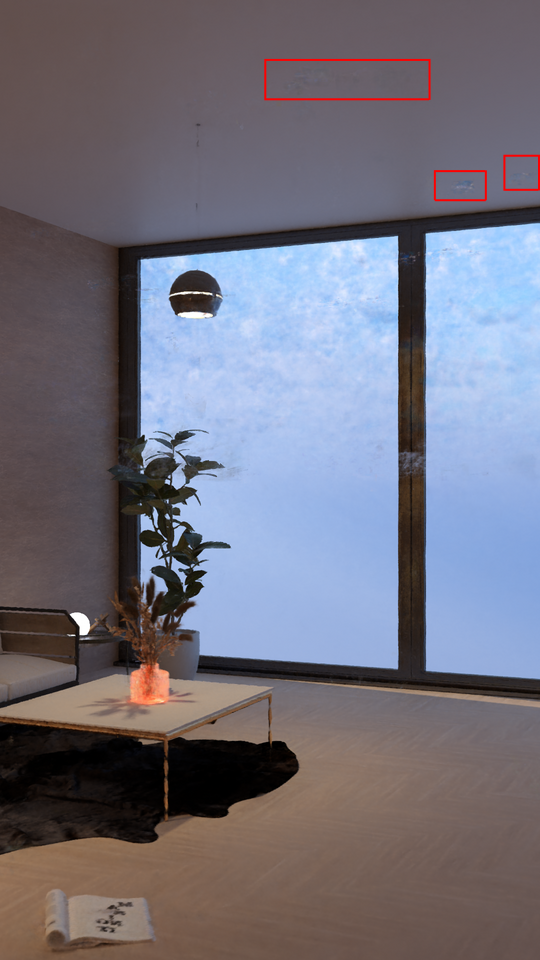}
        \subcaption{RGB}
    \end{minipage}%
    \begin{minipage}{0.12\textwidth}
        \centering
        \includegraphics[width=\linewidth]{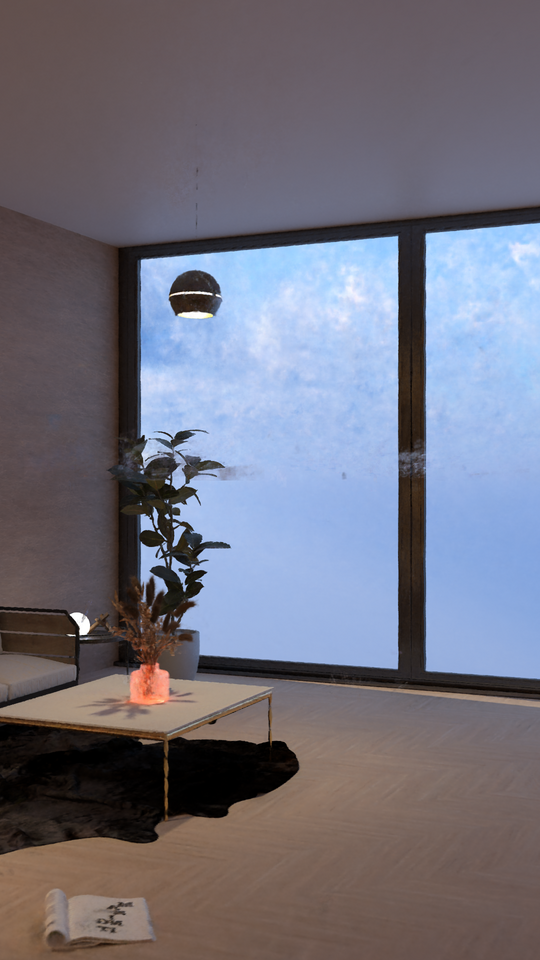}
        \subcaption{BoundL}
    \end{minipage}%
        \begin{minipage}{0.12\textwidth}
        \centering
        \includegraphics[width=\linewidth]{images/evaluation/livingroom/bad/c_424.png}
        \subcaption{RGB}
    \end{minipage}%
    \begin{minipage}{0.12\textwidth}
        \centering
        \includegraphics[width=\linewidth]{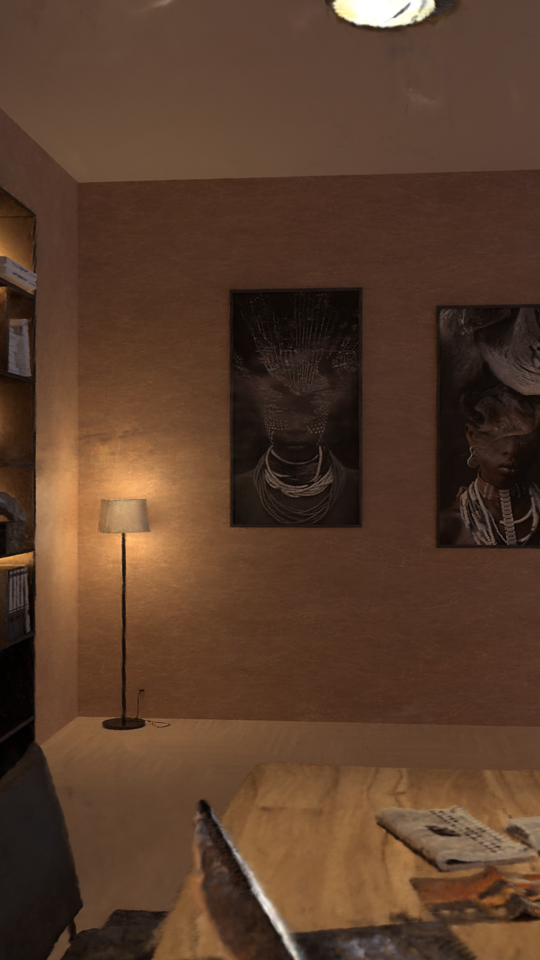}
        \subcaption{BoundL}
    \end{minipage}%
    \caption{Renderings with Instant-NGP trained on our 360-degree indoor dataset, using photometric loss in (a), (c), (e), and (g), and depth supervision with BoundL in (b), (d), (f), and (h). Red bounding boxes highlight floaters, which are minimized through depth guidance with planar architectural depth priors and BoundL.}
    \label{fig:rgb_vs_rgb_plus_depth}
\end{figure}

\begin{figure}[h!]
    \centering
    \captionsetup[subfigure]{labelformat=parens}
    \begin{minipage}{0.12\textwidth}
        \centering
        \includegraphics[width=\linewidth]{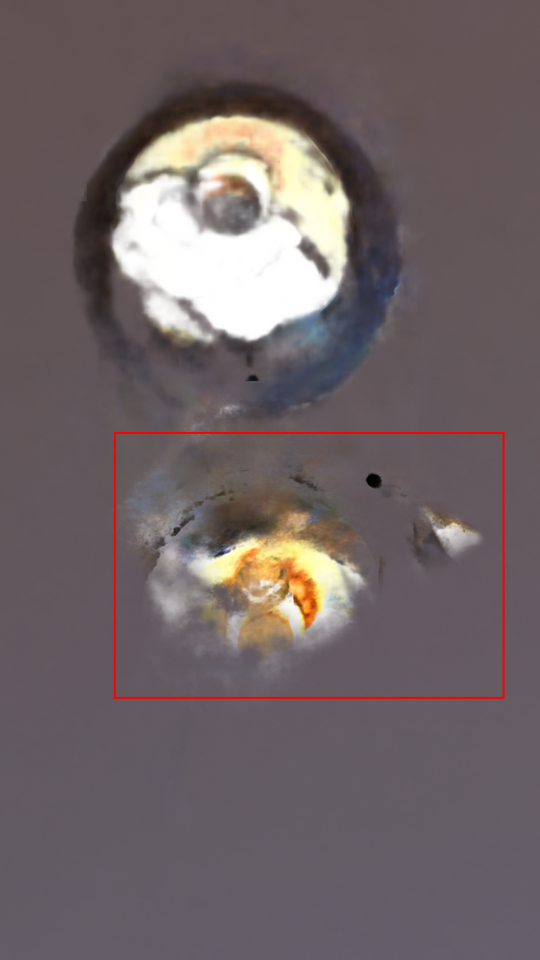}
        \subcaption{MSE}
    \end{minipage}%
    \begin{minipage}{0.12\textwidth}
        \centering
        \includegraphics[width=\linewidth]{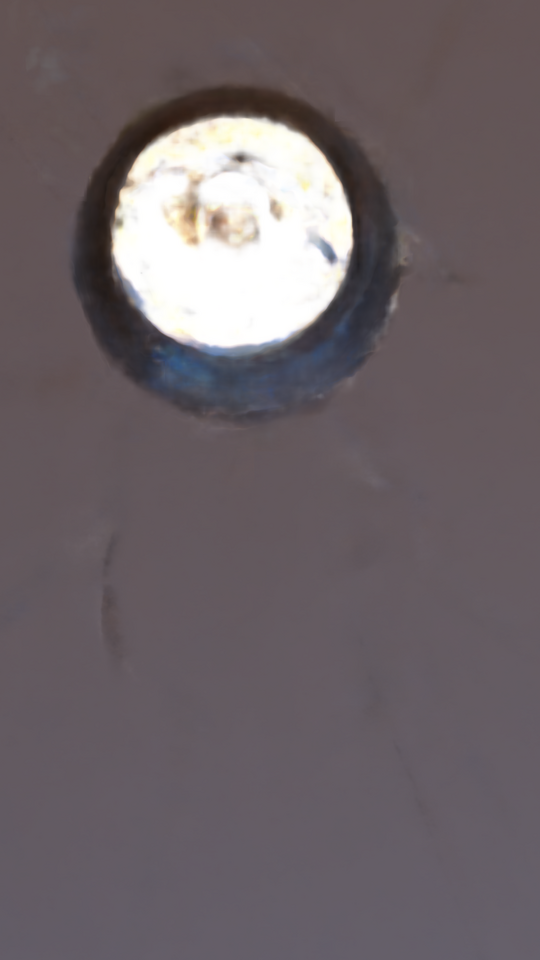}
        \subcaption{BoundL}
    \end{minipage}%
    \begin{minipage}{0.12\textwidth}
        \centering
        \includegraphics[width=\linewidth]{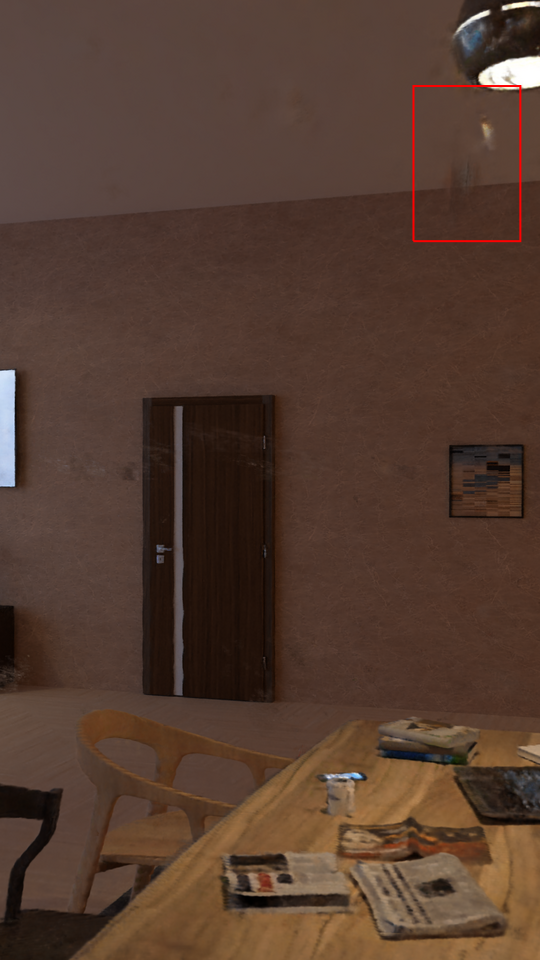}
        \subcaption{MSE}
    \end{minipage}%
    \begin{minipage}{0.12\textwidth}
        \centering
        \includegraphics[width=\linewidth]{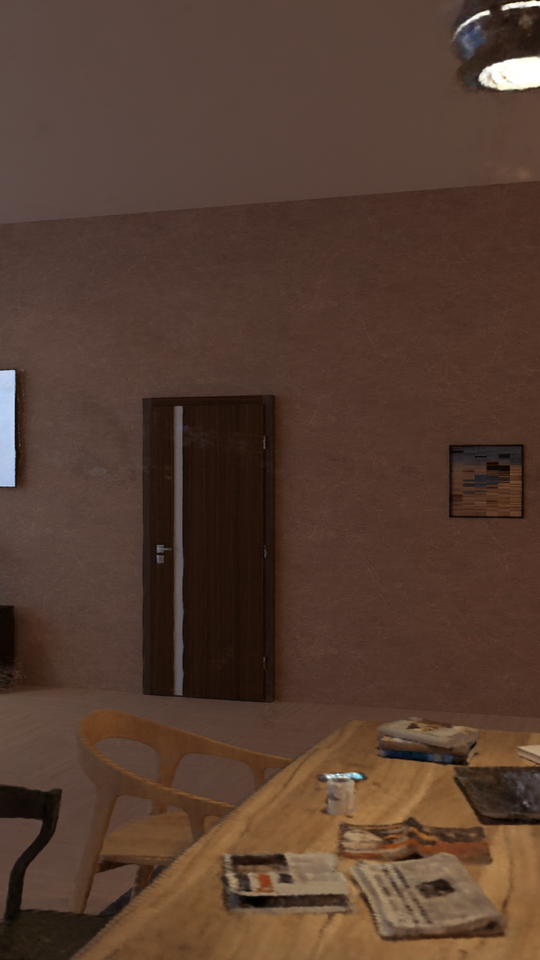}
        \subcaption{BoundL}
    \end{minipage}%
    
    \begin{minipage}{0.12\textwidth}
        \centering
        \includegraphics[width=\linewidth]{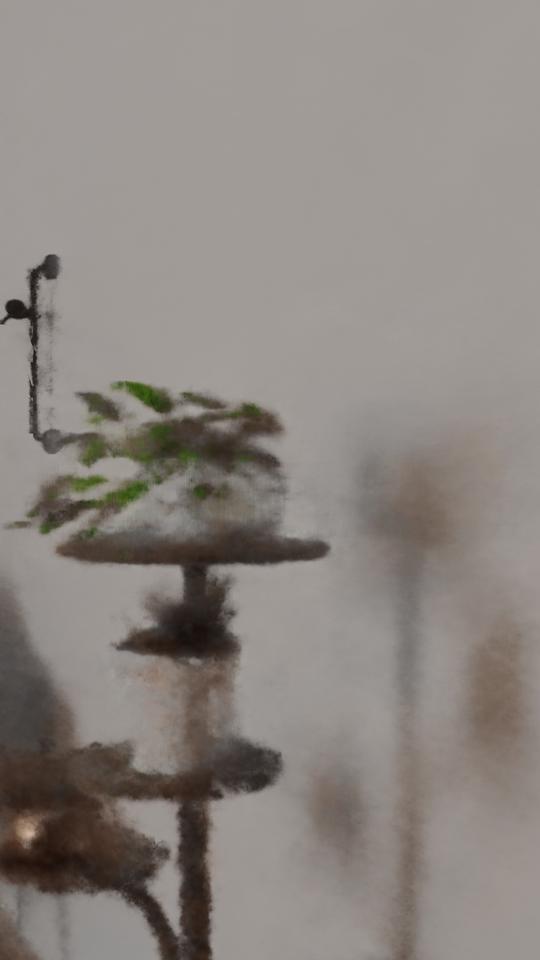}
        \subcaption{RGB}
    \end{minipage}%
    \begin{minipage}{0.12\textwidth}
        \centering
        \includegraphics[width=\linewidth]{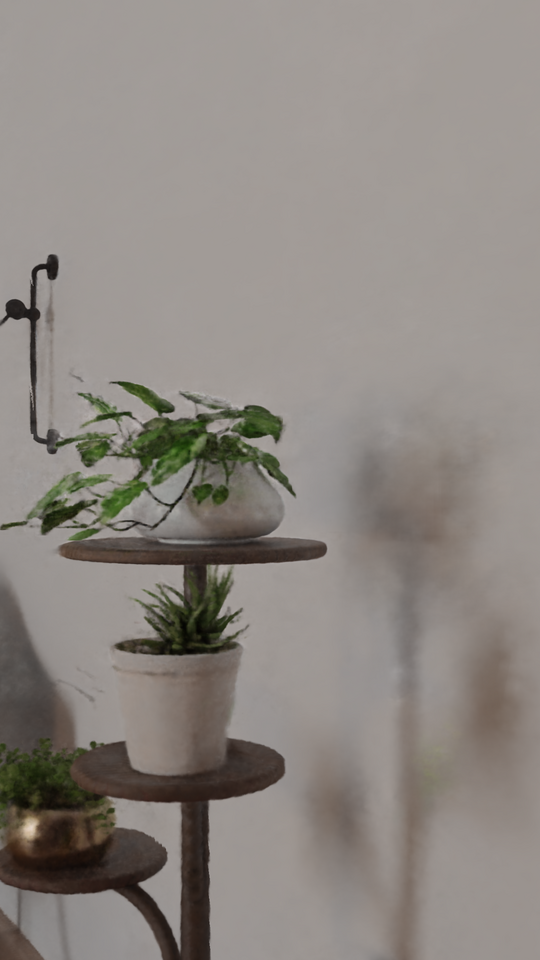}
        \subcaption{BoundL}
    \end{minipage}%
    \begin{minipage}{0.12\textwidth}
        \centering
        \includegraphics[width=\linewidth]{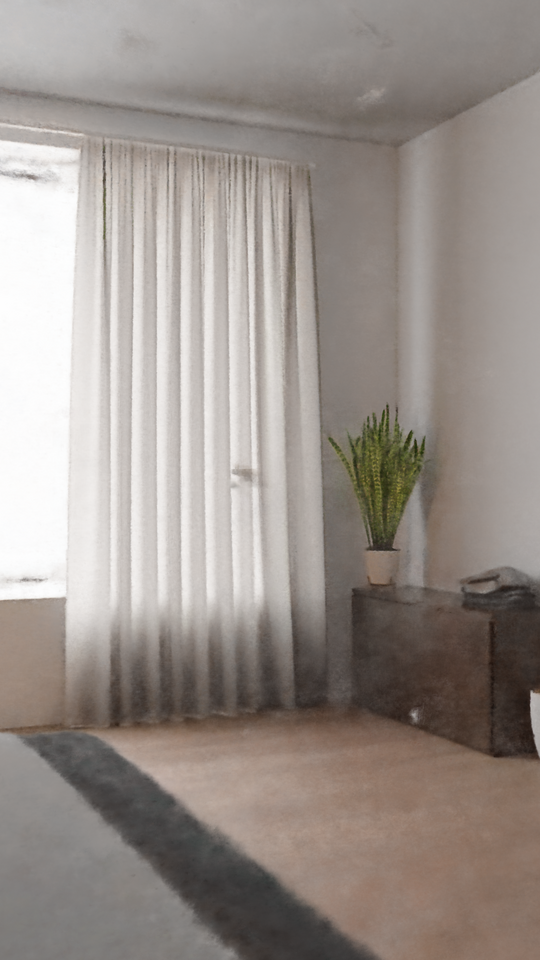}
        \subcaption{RGB}
    \end{minipage}%
    \begin{minipage}{0.12\textwidth}
        \centering
        \includegraphics[width=\linewidth]{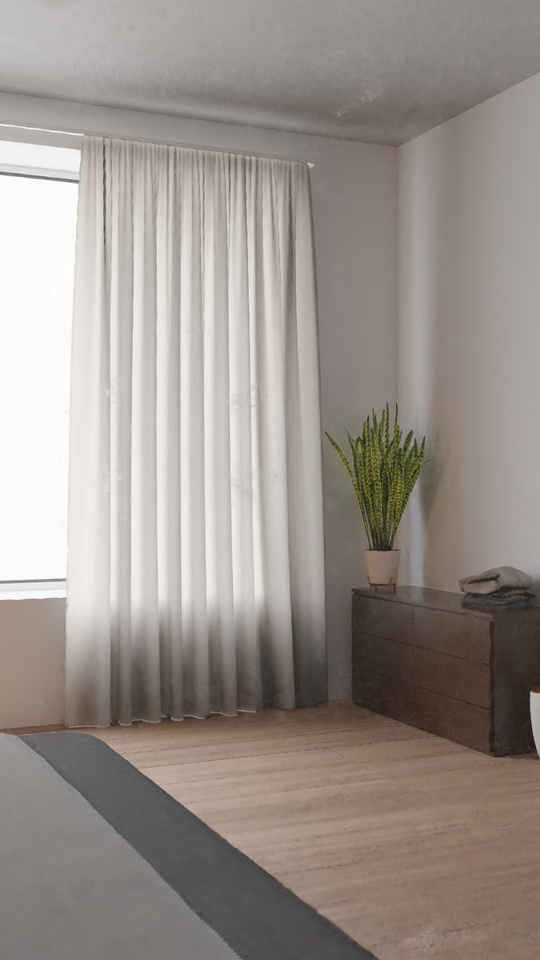}
        \subcaption{BoundL}
    \end{minipage}%
    \caption{Renderings with Instant-NGP trained on our 360-degree indoor dataset: (a)-(d) compare depth supervision methods, with MSE in (a) and (c), and BoundL in (b) and (d). Red boxes highlight areas where MSE results exhibit rendering artifacts that BoundL successfully mitigates. (e)-(h) illustrate the faster convergence of models with depth supervision, showing BoundL examples in (f) and (h) and photometric loss in (e) and (g).}
\label{fig:mse_vs_boundl_and_speed}
\end{figure}

Visual observations reveal a noticeable reduction in artifacts for depth-guided methods compared to those without depth supervision (see Figure \ref{fig:rgb_vs_rgb_plus_depth}). Moreover, the BoundL loss demonstrates fewer artifacts than MSE loss, producing cleaner and more accurate renderings (see Figure \ref{fig:mse_vs_boundl_and_speed}). This is likely due to BoundL's ability to directly address the weights of samples, effectively reducing ambiguity during the volume rendering procedure.

To quantitatively compare our models, we employ standard view synthesis evaluation metrics: Peak Signal-to-Noise Ratio (PSNR), Structural Similarity Index Measure (SSIM), and Learned Perceptual Image Patch Similarity (LPIPS). As expected, models with depth supervision outperform their counterparts, with BoundL loss (both with and without joint-bilateral regularization) achieving the highest metrics.

Our patch-based regularization methods deliver consistent quality improvements across both indoor scenes (see Table \ref{tab:final_results_comparison}), achieving better metrics compared to RegNeRF's depth patch regularization. This advantage is likely due to the ability of bilateral and joint-bilateral filtering to reduce noise while preserving sharp edge transitions and essential structural details. Moreover, joint-bilateral regularization demonstrates additional gains over the bilateral approach.

Notably, performance metrics vary across scenes, with the living room consistently outperforming the bedroom. This is likely due to obstructions in the bedroom—such as the large bed—limiting ray coverage in occluded areas.

Further, depth-supervised models also demonstrate faster convergence, requiring only 120,000 iterations compared to 200,000 for models trained solely on RGB loss (see Figure \ref{fig:mse_vs_boundl_and_speed}), and 280,000 iterations for those using patch-based depth supervision. This speedup is attributed to depth supervision, which enables the model to quickly identify empty spaces, concentrate sampling on occupied regions, and provide a stronger optimization signal \cite{deng2022depth}.

\begin{table*}[h]
\caption{Quantitative comparison for 360-degree indoor scenes on the evaluation dataset. We report PSNR, SSIM and LPIPS. ''Arch. planar'' refers to depth-guided methods that utilize depth priors for architectural planar surfaces.}

\centering
\begin{tabular}{|p{5cm}|c|c|c|c|c|c|}
\hline
\multirow{2}{*}{Method} & \multicolumn{3}{c|}{Bedroom Scene} & \multicolumn{3}{c|}{Livingroom Scene} \\ \cline{2-7}
 & PSNR $\uparrow$ & SSIM $\uparrow$ & LPIPS $\downarrow$ & PSNR $\uparrow$ & SSIM $\uparrow$ & LPIPS $\downarrow$ \\\hline
Only RGB loss & 31.163 & 0.749 & 0.378 & 34.008 & 0.856 & 0.268 \\\hline
Arch. planar + MSE & 34.174 & 0.762 & 0.327 & 36.790 & 0.869 & 0.252 \\\hline
Arch. planar + BoundL & 34.309 & \textbf{0.792} & 0.285 & 36.902 & \textbf{0.921} & 0.248 \\\hline
RegNeRF patch & 30.830 & 0.750 & 0.361 & 34.033 & 0.858 & 0.259 \\\hline
Bilateral filter & 31.890 & 0.750 & 0.359 & 35.200 & 0.858 & 0.256 \\\hline
Joint bilateral filter & 32.709 & 0.763 & 0.342 & 36.127 & 0.871 & 0.256 \\\hline
Arch. planar + BoundL with joint bilateral & \textbf{34.410} & 0.765 & \textbf{0.281} & \textbf{36.935} & 0.898 & \textbf{0.245} \\\hline
\end{tabular}
\label{tab:final_results_comparison}
\end{table*}

\section{\uppercase{Conclusions}}
\label{sec:conclusion}
This research tackles the challenge of textureless regions for NeRF-based novel view synthesis in indoor environments. To address this, we developed a depth guidance approach for large planar surfaces, such as walls, floors, and ceilings—regions where NeRFs often struggle. Specifically, we proposed an efficient method to compute depth priors for the mentioned surfaces and
introduced a depth loss function, BoundL, to enforce depth constraints on these planar boundaries. This is complemented by our patch-based regularization, which utilizes bilateral and joint-bilateral filtering.

To evaluate our approach, we created a synthetic indoor dataset comprising two distinct scenes that simulate individual views within a 360-degree panorama prior to assembly. Working with a series of raw images captured with a pinhole camera model aids in determining accurate image poses, eliminating the need to account for geometric distortions in the final 360-degree stitched panorama.

Our results demonstrate clear improvements in rendering quality, both visually and quantitatively, when incorporating our planar depth priors with depth supervision through MSE and BoundL loss. Notably, BoundL consistently outperforms MSE across both scenes. Additionally, our patch regularization techniques surpass RegNeRF’s patch depth constraints, yielding subtle yet stable quantitative gains.

With all enhancements enabled, we achieved an increase in PSNR of up to 3 dB compared to the baseline model using only photometric loss. These improvements underscore the robustness and effectiveness of our approach in refining NeRF rendering for complex indoor environments.

Future work will extend our methods to real-world data, with optimizations to account for noisy camera parameters. Additionally, incorporating sparse depth data from feature-rich regions and enforcing strict planarity on other linear surfaces could further improve model accuracy and rendering quality.

\section*{\uppercase{Acknowledgements}}
This work has partly been funded by the German Federal Ministry for
Digital and Transport (project EConoM, grant
no~19OI22009C) and the German Federal Ministry of Education and Research (project VoluProf, grant no.~16SV8705).



\bibliographystyle{apalike}
{\small
\bibliography{citations}}

\end{document}